\pdfoutput=1 

\newif\REVIEW

\ifdefined\REVIEW
    \documentclass[
        a4paper,
        onecolumn
    ]{article}
    \usepackage{lineno, setspace}
\else
    \documentclass[
        a4paper,
        twocolumn
    ]{article}
\fi

\usepackage{IEK10} 
\usepackage{natbib}

\usepackage{upgreek}
\usepackage{bbm}

\DeclareSIUnit\year{yr}
\DeclareSIUnit\annum{a}
\DeclareSIUnit\ton{t}
\DeclareSIUnit{\million}{\text{M}}

\def\IEK10{
  Institute of Climate and Energy Systems,
  Energy Systems Engineering (ICE-1),
  Forschungszentrum J\"ulich GmbH,
  J\"ulich 52425,
  Germany
}
\def\RWTH{
  RWTH Aachen University,
  Aachen 52062,
  Germany
}
\def\JARA{
  JARA-ENERGY,
  J{\"u}lich 52425,
  Germany
}
\def\SVT{
  RWTH Aachen University,
  Process Systems Engineering (AVT.SVT),
  Aachen 52074,
  Germany
}

\def\UCL{
  Department of Chemical Engineering, 
  Sargent Centre for Process Systems Engineering, 
  University College London (UCL), 
  London, WC1E 7JE, 
  United Kingdom
}

\newcommand{\mytitle}{Data-Driven Conditional Flexibility Index}

\newcommand{\affil}{
  \begin{itemize}[leftmargin=3mm, itemsep=0mm]
    \item[$^a$]\IEK10
    \item[$^b$]\RWTH
    \item[$^c$]\UCL
    \item[$^d$]\SVT
    \item[$^e$]\JARA
  \end{itemize}
}

\def\firstAuthor{Moritz Wedemeyer}
\newcommand{\myauthor}{\firstAuthor$^{a,b}$,
Eike Cramer$^{c,d}$,
Alexander Mitsos$^{e,a,d}$,
Manuel Dahmen$^{a,*}$
}
\newcommand{\correspondingauthor}{M. Dahmen, \IEK10 \newline E-mail: m.dahmen@fz-juelich.de}
\author{\myauthor}

\usepackage[
  colorlinks,
  linkcolor=blue,
  citecolor=blue,
  urlcolor=blue,
  pdftitle={\mytitle},
  pdfauthor={\firstAuthor}
]{hyperref}
\usepackage[capitalise, nameinlink]{cleveref}
\crefname{table}{Tab.}{Tab.}

\usepackage[colorinlistoftodos]{todonotes}

\begin{document}
\twocolumn[
\begin{@twocolumnfalse}

  \thispagestyle{firststyle}

  \begin{center}
    \begin{large}
      \textbf{\mytitle}
    \end{large} \\
    \myauthor
  \end{center}

  \vspace{0.5cm}

  \begin{footnotesize}
    \affil
  \end{footnotesize}

  \vspace{0.5cm}
    With the increasing flexibilization of processes, determining robust scheduling decisions has become an important goal. 
    Traditionally, the flexibility index has been used to identify safe operating schedules by approximating the admissible uncertainty region using simple admissible uncertainty sets, such as hypercubes.
    Presently, available contextual information, such as forecasts, has not been considered to define the admissible uncertainty set when
    determining the flexibility index.
    We propose the \emph{conditional flexibility index} (CFI), which extends the traditional flexibility index in two ways: by learning the parametrized admissible uncertainty set from historical data and by using contextual information to make the admissible uncertainty set conditional.
    This is achieved using a normalizing flow that learns a bijective mapping from a Gaussian base distribution to the data distribution.
    The admissible latent uncertainty set is constructed as a hypersphere in the latent space and mapped to the data space.
    By incorporating contextual information, the CFI provides a more informative estimate of flexibility by defining admissible uncertainty sets in regions that are more likely to be relevant under given conditions.
    Using an illustrative example, we show that no general statement can be made about data-driven admissible uncertainty sets outperforming simple sets, or conditional sets outperforming unconditional ones.
    However, both data-driven and conditional admissible uncertainty sets ensure that only regions of the uncertain parameter space containing realizations are considered.
    We apply the CFI to a security-constrained unit commitment example and demonstrate that the CFI can improve scheduling quality by incorporating temporal information.
  \vspace{0.5cm}


\end{@twocolumnfalse}
]
\ifdefined\REVIEW
  \onecolumn
\else
  \newpage
\fi

\section{Introduction}\label{sec:Intro}
    Accounting for uncertainty when determining scheduling decisions in complex operational settings, such as the optimization of power systems, is an important task.
    When safety aspects are affected, for example, to ensure that transmission lines are not overloaded, robust optimization \citep{soysterTechnicalNoteConvex1973, ben-talrobustoptimization2009} can be applied to guarantee constraint satisfaction under uncertainty.
    An important formulation, which can be classified under the robust optimization paradigm, is the flexibility index, originally introduced by \cite{swaneyIndexOperationalFlexibility1985}, as the ability of a system to operate feasibly under different operating conditions.
    While the flexibility index was initially developed to quantify operational feasibility under uncertainty for steady-state processes, it has since been extended to dynamic systems by \cite{dimitriadisFlexibilityAnalysisDynamic1995}.
    The growing availability of high-resolution data, such as weather observations and forecasts, raises the question of how such information can be exploited to obtain improved estimates of operational flexibility.

    Within the flexibility index context, three major tasks can be differentiated:
    (i) the flexibility test problem \citep{grossmannOptimizationStrategiesFlexible1983}, which determines whether a given system can operate feasibly for any uncertainty realization within a predefined uncertainty set.
    (ii) the flexibility index problem \citep{swaneyIndexOperationalFlexibility1985}, which is concerned with identifying an index that characterizes the size of the region of uncertain parameters in which a system can feasibly operate.
    (iii) the flexibility index maximization problem, where the flexibility index is maximized as part of the objective function by adjusting design or scheduling decisions of a system \citep{halemaneoptimalprocessdesign1983, grossmannEvolutionConceptsModels2014}.

\begin{figure}
        \centering
        \includegraphics[width=\columnwidth]{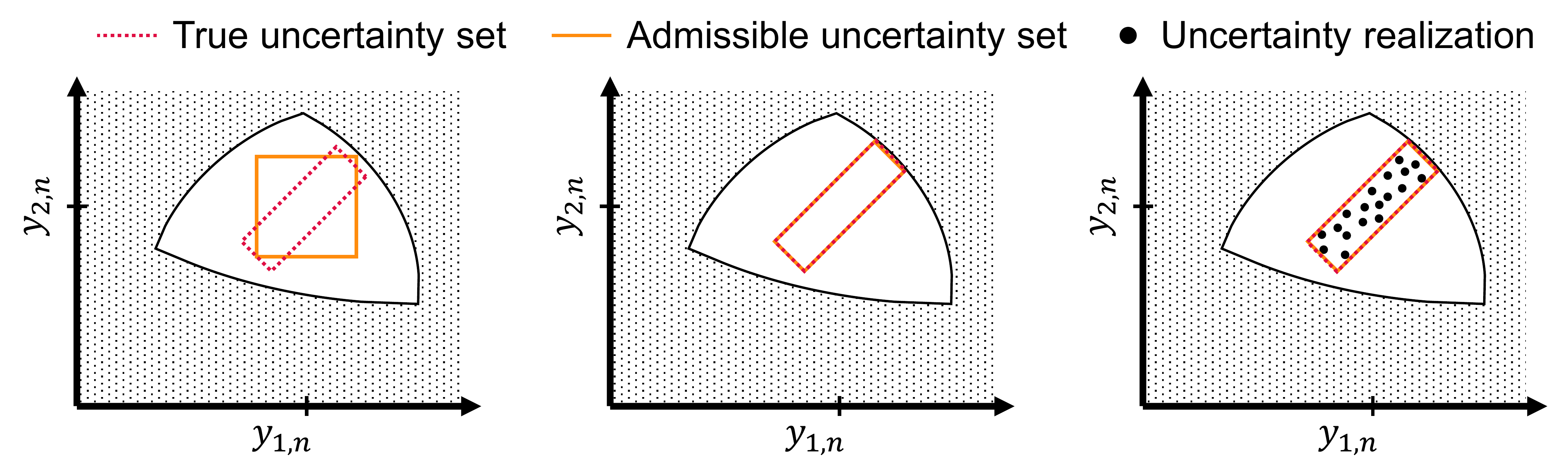}
        \caption{Illustration of the flexibility index problem. The white area represents the admissible uncertainty region, while shaded areas indicate infeasible parameter realizations. Nominal parameter realizations $y_{i,n}$ are marked. The true uncertainty set comprises all possible uncertain parameter realizations and is unknown in practice; it is shown here only for visualization.
        The left panel illustrates the traditional approach of approximating the admissible uncertainty region with a hypercube without any knowledge of the underlying true uncertainty set. The admissible uncertainty set contains regions where no uncertainty realizations occur. The middle panel shows improved approximation when correct parameter correlations are assumed; in this special case, the admissible uncertainty set coincides with the true uncertainty set, which is not generally the case. The right panel shows uncertain parameter realizations (black dots) used to estimate parameter correlations.}
        \label{fig:FlexInd}
    \end{figure}

    In the flexibility index context, the set of uncertain parameter values that permit feasible operation is commonly referred to as the \emph{feasible region} (see, e.g., \cite{swaneyIndexOperationalFlexibility1985, grossmannEvolutionConceptsModels2014}).
    To avoid confusion with the feasible region in the space of the design variables, the feasible region in the space of the uncertain parameters is hereafter referred to as the \emph{admissible uncertainty region}.
    When determining the flexibility index, the size of the admissible uncertainty region is approximated using an \emph{admissible uncertainty set}.    
    This set is parametrized by the flexibility index and may vary in size, the term \emph{admissible} is introduced to distinguish it from fixed-size uncertainty sets typically used in robust optimization.
    Because the admissible uncertainty region can have a complex shape, its size is usually approximated by using a simple admissible uncertainty set, centered at the expected nominal value of the uncertain parameters.
    Consequently, the choice of the nominal parameter values directly affects the size and location of the admissible uncertainty set.
    A common choice is a scaled hyper-rectangle, centered at the nominal parameter realization, where the flexibility index determines the side lengths \citep{swaneyIndexOperationalFlexibility1985, grossmannEvolutionConceptsModels2014}, which is shown at the left of Figure \ref{fig:FlexInd}.
    However, Figure \ref{fig:FlexInd} also showcases that simple admissible uncertainty sets may substantially underestimate the true admissible uncertainty region and may include areas that do not contain any uncertain parameter realizations.
    To improve the accuracy of the flexibility index and the true flexibility achieved by the flexibility index maximization problem, i.e., the region of uncertain parameters in which a process can operate feasibly, efforts to better approximate the shape of the admissible uncertainty region have been undertaken.
    \cite{ierapetritouNewApproachQuantifying2001} and \cite{goyalDeterminationOperabilityLimits2002} approximated the admissible uncertainty region by a convex polytope.
    However, their approach only allows the approximation of convex admissible uncertainty regions.
    \cite{kucherenkoAnalyticalIdentificationProcess2025} used R-functions and polynomial approximations of individual constraints to identify the shape of the admissible uncertainty region.

    The approximation methods discussed above do not account for potential correlations between uncertain parameters.
    The difference between the traditional flexibility index and an approach accounting for parameter correlations determined from \emph{historical data} is illustrated in Figure \ref{fig:FlexInd}.
    By learning the correlations in the historical data and embedding them in the flexibility index formulation, one can obtain admissible uncertainty sets that better align with realistic realizations, leading to more accurate flexibility estimates.
    Since \cite{grossmannActiveConstraintStrategy1987} introduced the first approach able to handle correlated uncertain parameters, multiple ways to account for correlation have been proposed:
    \cite{rooneyDesignModelParameter2001} used a nonconvex admissible uncertainty set derived from the likelihood ratio test, while       \cite{langnerFlexibilityAnalysisUsing2023} used hyperpolygons.
    \cite{pulsipherMixedintegerConicProgramming2018} considered correlations of uncertain parameters by transferring the use of ellipsoidal admissible uncertainty sets introduced by \cite{rooneyIncorporatingJointConfidence1999} to the flexibility index problem.
    Their result provides a lower bound to the so-called stochastic flexibility index \citep{pistikopoulosNovelFlexibilityAnalysis1990, straubIntegratedStochasticMetric1990, straubDesignOptimizationStochastic1993}, which, given a joint probability distribution of the uncertain parameters, guarantees that the process will operate feasibly with a certain probability.

    In the search for better approximations of the shape of the admissible uncertainty region, machine learning methods have seen wide application.
    It is important to note that the methods mentioned in this paragraph do not make use of historical data of uncertain parameters but instead try to find a good approximation of the admissible uncertainty region, i.e., the white area in Figure \ref{fig:FlexInd}.
    \cite{boukouvalaFeasibilityAnalysisBlackbox2012} and    \cite{geremiaNovelFrameworkIdentification2023} used Gaussian processes to approximate the admissible uncertainty region, while    \cite{mettaNovelAdaptiveSampling2021} used neural networks.
    \cite{zhangNovelFeasibleSet2024, zhangDynamicProcessFlexibility2025} trained a convolutional neural network classifier that maps combinations of uncertain parameters into a feature space.
    The classifier was trained to map feasible uncertain parameter combinations close to a point while maximizing the distance of infeasible points from that point.
    The flexibility index can then be defined as the ratio of random uncertain parameter samples that are mapped within a hypersphere with a radius equal to the distance of the furthest training point from the center of the sphere.

    The concept of the flexibility index is closely related to the field of robust optimization \citep{ben-talrobustoptimization2009, zhangRelationFlexibilityAnalysis2016}, which emerged more recently.
    Similar to the developments in the flexibility index literature, several authors have proposed to incorporate historical data into the construction of uncertainty sets for robust optimization, introducing the field of data-driven robust optimization \citep{bertsimasDatadrivenRobustOptimization2018}.
    \cite{hongLearningBasedRobustOptimization2020} learned the shape of the uncertainty set from data using simple shapes such as ellipsoids and polytopes.
    They calibrated the size of the uncertainty set to achieve a desired feasibility probability by using a second part of the dataset that was held out during the learning of the shape.
    \cite{goerigkDatadrivenRobustOptimization2023} trained neural network classifiers to be embedded into the robust optimization problem as constraints defining the uncertainty set.
    \cite{dumouchelle2024neurro} trained a neural network to identify worst-case scenarios for two-stage robust optimization, replacing an embedded maxmin problem with a neural network approximation.
    Two-stage robust optimization includes two sets of decision variables, where variables in the first set have to be fixed before the uncertainty is realized.
    In contrast, variables in the second set can be adjusted after the uncertainty is realized.
    \cite{brennerDeepGenerativeLearning2025} constructed the uncertainty set by using a variational autoencoder to learn the mapping between a latent space and the uncertainty space, which is then used to transform a simple hyper-sphere uncertainty set from the latent space into the data space.
    \cite{tulabandhulaRobustOptimizationUsing2014} introduced a method to construct uncertainty sets for robust optimization from machine learning models.
    
    \cite{tulabandhulaRobustOptimizationUsing2014} already introduced an important concept that only gained significant traction very recently: The machine learning model incorporates \emph{forecast information}, leading to uncertainty sets that are conditioned on contextual information.
    \cite{NEURIPS2022_3df87436} later referred to this concept as \emph{conditional} data-driven robust optimization.
    They built on \cite{goerigkDatadrivenRobustOptimization2023} by clustering the contextual information and training a dedicated neural network classifier for each cluster.
    Going further, \cite{chenreddyEndtoendConditionalRobust2024} introduced an end-to-end learning approach where they find task-optimal conditional ellipsoidal uncertainty sets.
    Instead of fitting the ellipsoidal uncertainty sets to the data, they take into account their performance in the downstream optimization task.
    Conditional robust optimization has been applied to power systems by \cite{wanIntegratedProbabilisticForecasting2025}, who incorporated forecasting information into the robust optimal power flow problem by training a model to predict the bounds for a box-constrained uncertainty set based on forecasting information.

    Building on these foundations, we introduce the \emph{conditional flexibility index} (CFI),
    which brings conditional, data-driven admissible uncertainty sets into the flexibility index framework and aims to measure the size of the admissible uncertainty region of uncertain parameters given contextual information.
    Unlike prior machine-learning-based work in this area, e.g., \cite{boukouvalaFeasibilityAnalysisBlackbox2012} and \cite{zhangDynamicProcessFlexibility2025}, our approach learns admissible uncertainty sets that approximate the shape of the support of the distribution of \emph{uncertain parameters}, not the admissible uncertainty region.
    Moreover, by using realizations of the uncertain parameters to learn the admissible uncertainty set, we can include contextual information to learn conditional admissible uncertainty sets.
    This is particularly advantageous in settings such as power systems optimization, where safe operation is critical and both historical data of uncertain parameter realizations and forecast information are available, enabling the learning of complex conditional admissible uncertainty sets.
    By conditioning on relevant information, the CFI restricts attention to uncertainty realizations that are likely to occur, leading to scheduling decisions that are robust with respect to the actual expected realizations.
    To determine the CFI, we propose using \emph{normalizing flows} to learn the parametrized conditional admissible uncertainty set from historical data.        
    Different from the work of \cite{NEURIPS2022_3df87436}, who cluster contextual information and train separate machine learning models for each cluster, we train a single normalizing flow model that can take continuous contextual information as input to construct conditional admissible uncertainty sets.
   
    We embed the normalizing flow into the flexibility index problem and compute the flexibility index as the radius of a hypersphere in the latent space of the normalizing flow, leading to a conditional robust optimization problem.
    The construction of our admissible uncertainty set is similar to the approach by \cite{brennerDeepGenerativeLearning2025}, but our method  additionally incorporates conditional information into the construction of the admissible uncertainty sets.
    Whereas \cite{brennerDeepGenerativeLearning2025} employ a gradient-ascent procedure that differentiates through the recourse problem and therefore requires convexity with respect to the recourse actions~\citep{amos}, we instead formulate the conditional robust optimization problem as a semi-infinite program and solve it using an adaptive discretization scheme,    allowing us to handle any two-stage robust optimization problem.

    The remainder of this work is structured as follows: Section \ref{sec:Method} introduces the theoretical background and solution approaches.
    Section \ref{sec:Illustrative_Example} demonstrates our proposed CFI approach on an illustrative example with binary contextual information, while also highlighting some limitations of the learnable admissible uncertainty sets.
    Section \ref{sec:Case_Study} applies the CFI approach to a security-constrained unit commitment (SCUC) problem under nodal power injection uncertainty with continuous contextual information.
    Finally, Section \ref{sec:conclusion} concludes our work.
\section{Methods} \label{sec:Method}
    In the following section, we introduce the formulation of maximizing the conditional flexibility index.
    Furthermore, we provide a brief overview of normalizing flows and their properties, and describe our approach to embedding them into the optimization problem.
    \subsection{The flexibility index maximization problem}\label{sec:FIP}
    Formally, the problem of maximizing the flexibility index $\delta$ can be written as
    \begin{equation}
    \begin{aligned}\label{flex_ind}
        \delta^{*} = &\underset{\delta \in \mathbb{R}, \mathbf{x} \in \mathcal{X}}\max \quad \delta & \\
        &\text{s.t.} \quad  \underset{\mathbf{y} \in \mathcal{Y}(\delta)}\max\underset{\mathbf{z} \in \mathcal{Z}(\mathbf{x}, \mathbf{y})}\min \quad g(\mathbf{x}, \mathbf{y}, \mathbf{z}) \le 0,
    \end{aligned}
    \end{equation}
    where $\mathbf{x}$ denotes the design decisions, $\mathbf{y}$ the uncertainty realizations, and $\mathbf{z}$ the operational decisions that can be taken in response to the uncertainty realization.
    The corresponding feasible sets are denoted by $\mathcal{X}$, $\mathcal{Y}(\delta)$, and $\mathcal{Z}(\mathbf{x}, \mathbf{y})$, respectively.
    $g(\mathbf{x}, \mathbf{y}, \mathbf{z})$ is the constraint that needs to be satisfied to ensure feasible operation.
    Vector-valued constraints can be handled by defining
    \begin{equation*}
    g(\mathbf{x}, \mathbf{y}, \mathbf{z}) = \underset{j \in \mathcal{J}}\max \ g_{j}(\mathbf{x}, \mathbf{y}, \mathbf{z}),
    \end{equation*}
    where $g(\mathbf{x}, \mathbf{y}, \mathbf{z})$ represents the maximum value among the components $g_{j}(\mathbf{x}, \mathbf{y}, \mathbf{z})$ of the vector-valued constraint indexed by $j \in \mathcal{J}$. 
    Solution strategies have been mainly developed for the flexibility index problem where the design variables $\mathbf{x}$ are fixed.
    \cite{swaneyIndexOperationalFlexibility1985} showed that for special cases, vertex enumeration can be used to solve \eqref{flex_ind}.
    Furthermore, \cite{grossmannActiveConstraintStrategy1987} introduced an active set method to solve the flexibility index problem, enabling the handling of correlated uncertain parameters \citep{grossmannEvolutionConceptsModels2014}.
    Similarly, \cite{zhaoNovelFormulationsFlexibility2022} replace the lower-level problem by its KKT conditions \citep{karushMinimaFunctionsSeveral2014, kuhnNonlinearProgramming1951} to solve the flexibility index problem.
    Both the active set and KKT-condition approach can only guarantee a global solution for convex model equations \citep{grossmannEvolutionConceptsModels2014}; in the general nonconvex case, the identified flexibility index might be erroneous.
    Problem \eqref{flex_ind} is an existence-constrained semi-infinite program (ESIP) \citep{djelassiglobalsolutionsemiinfinite2021}.
    Unlike regular semi-infinite programs (SIPs) \citep{charnesdualityhaarprograms1962, hettichsemiinfiniteprogrammingtheory1993, steinhowsolvesemiinfinite2012} the ESIP has recourse variables $\mathbf{z}$ which can be adjusted after the uncertainty has realized.
    Furthermore, because the set of uncertainty realizations $\mathcal{Y}(\delta)$ is parametrized by $\delta$, Problem \eqref{flex_ind} is a generalized existence-constrained semi-infinite programming (EGSIP) problem \citep{djelassidiskretisierungsbasiertealgorithmenfur2020} which analogously to generalized semi-infinite programs (GSIPs) \citep{ steinGeneralizedSemiinfiniteOptimization2002, guerravazquezgeneralizedsemiinfiniteprogramming2008, mitsosglobaloptimizationgeneralized2015} can be solved via a discretization based solution approach \citep{djelassiglobalsolutionsemiinfinite2021}, assuming the ESIP relaxation is valid.
    
    The following problem formulation describes the maximization of the \emph{conditional} flexibility index:
    \begin{equation}
    \begin{aligned}\label{cond_flex_ind}
        \delta_{cond}^{*}(\mathbf{c}) = &\underset{\delta \in \mathbb{R}, \mathbf{x} \in \mathcal{X}}\max \quad \delta & \\
        &\text{s.t.} \quad  \underset{\mathbf{y} \in \mathcal{Y}_{cond}(\mathbf{c}, \delta)}\max\underset{\mathbf{z} \in \mathcal{Z}(\mathbf{x}, \mathbf{y})}\min \quad g(\mathbf{x}, \mathbf{y}, \mathbf{z}) \le 0
    \end{aligned}
    \end{equation}
    The critical difference between Problem \eqref{flex_ind} and Problem \eqref{cond_flex_ind} is that the admissible uncertainty set $\mathcal{Y}_{cond}(\mathbf{c}, \delta)$ depends on contextual information $\mathbf{c}$, leading to a parametric program \citep{manneNotesParametricLinear1953, OBERDIECK201661}.
    In our setting, however, we do not solve \eqref{cond_flex_ind} parametrically, instead we fix the contextual information $\mathbf{c}$ to a given value and solve the resulting instance.
    Similar to Problem \eqref{flex_ind}, Problem \eqref{cond_flex_ind} is an EGSIP and can be solved via a discretization-based solution approach \citep{djelassiglobalsolutionsemiinfinite2021}.
    The important question to formulate Problem \eqref{cond_flex_ind}  is: How can the conditional admissible uncertainty set $\mathcal{Y}(\mathbf{c}, \delta)$ be constructed?

    As introduced in Section \ref{sec:Intro}, machine learning models have been used to define uncertainty sets for robust optimization.
    However, unlike in robust optimization, the size of the admissible uncertainty set in flexibility index maximization depends on the flexibility index $\delta$; hence, the machine learning model needs to be able to define varying sizes of admissible uncertainty sets parametrized by the flexibility index $\delta$.

    \subsection{Normalizing flow-based admissible uncertainty sets}\label{sec:NFUC}
    We propose to use conditional normalizing flows to construct the admissible uncertainty set.
    A normalizing flow is a generative machine learning model that enables sampling and density estimation of complex data distributions \citep{kobyzevNormalizingFlowsIntroduction2021, papamakarios2021normalizing}.
    Normalizing flows are composed of a diffeomorphic, i.e., bijective and differentiable, transformation $\mathbf{f}$ and a simple base distribution $p_{l}(\mathbf{l})$, typically, a Gaussian distribution.
    Samples from the data distribution $p_{y}(\mathbf{y})$ are obtained by drawing from the base distribution $\mathbf{l} \sim p_{l}(\mathbf{l})$ and applying the transformation
    \begin{equation*}
    \mathbf{y} = \mathbf{f}(\mathbf{l}).
    \end{equation*}
    Because $\mathbf{f}$ is diffeomorphic, the probability density function of the data follows from the change-of-variables formula \citep{papamakarios2021normalizing}
     \begin{equation*}
          p_{y}(\mathbf{y}) = p_{l}(\mathbf{l}) \left|\det{J_{\mathbf{f}}(\mathbf{l})}\right|^{-1},
     \end{equation*}
     which, expressed in terms of the inverse transformation $\mathbf{l} = \mathbf{f}^{-1}(\mathbf{y})$, becomes
    \begin{equation}\label{eq:flowobj}
          p_{y}(\mathbf{y}) = p_{l}(\mathbf{f}^{-1}(\mathbf{y})) \left|\det{J_{\mathbf{f}^{-1}}(\mathbf{y})}\right|,
     \end{equation}
    where $\det{J_{\mathbf{f}}(\mathbf{l})}$ is the determinant of the Jacobian of $\mathbf{f}$ and $\det{J_{\mathbf{f}^{-1}}(\mathbf{y})}$ is the determinant of the Jacobian of the inverse transformation $\mathbf{f}^{-1}$.
    Equation \ref{eq:flowobj} can then be used to train the normalizing flow via log-likelihood maximization \citep{papamakarios2021normalizing}.

    Furthermore, normalizing flows can also be used to learn conditional probability distributions \citep{winkler2019learning}
    \begin{equation*}
      p_{y | c}(\mathbf{y} | \mathbf{c}) = p_{l}(\mathbf{l}) \left|\det{J_{\mathbf{f}}(\mathbf{l}, \mathbf{c})}\right|^{-1},
    \end{equation*}
    where the transformation $\mathbf{f}(\cdot, \mathbf{c})$ depends on contextual information $\mathbf{c}$.

    To construct conditional admissible uncertainty sets, we train a conditional normalizing flow on historical realizations of the uncertain parameters and their associated contextual information.
    In the traditional flexibility index framework, the expected values and deviations of the uncertain parameters are typically specified by the user based on experience or statistical data \citep{swaneyIndexOperationalFlexibility1985}.
    In contrast, our approach does not require any such specification: During normalizing flow training, both the shape and the center of the admissible uncertainty set are learned directly from data, and the learned center may differ from the expected value of the uncertain parameters.
    However, because the center of the admissible uncertainty set is not fixed a priori in the CFI approach and directly influences the resulting flexibility index, comparison with established methods becomes difficult unless differences between the centers are properly accounted for.
    We define the admissible uncertainty set as a hypersphere  in the latent space, i.e., the space defined by the support of the base distribution, of the normalizing flow $\mathcal{L} = \left\{ \mathbf{l} \ | \ \left\lVert \mathbf{l} \right\rVert_{2}^2 - \delta \le 0 \right \}$ and map it through the normalizing flow to obtain the corresponding admissible uncertainty set in the data space,
    \begin{equation*}
     \mathcal{Y}(\mathbf{c}) = \mathbf{f}(\mathcal{L}, \mathbf{c}) = \left\{ \mathbf{f}(\mathbf{l}, \mathbf{c}) \ | \ \mathbf{l} \in \mathcal{L} \right \}.
    \end{equation*}
    The flexibility index $\delta$ is defined as the squared radius of the latent-space hypersphere; the latent space admissible uncertainty set is the quadratic constraint above; the objective function is linear; overall, avoiding the introduction of a square root.
    Our approach is similar to that of \cite{brennerDeepGenerativeLearning2025}, which defines the uncertainty set as a hypersphere in the latent space of a variational autoencoder to construct unconditional uncertainty sets.

    Under the standard Gaussian base distribution, the latent-space density decays with increasing radius, and the coverage probability of $\mathcal{L}$ is known analytically.
    For a $k$-dimensional multivariate unit Gaussian, the analytical coverage probability is equal to the value of the cumulative distribution function of a $\chi^{2}$-distribution with $k$ degrees of freedom, evaluated at $\chi^{2} = \delta$ \citep{slotaniToleranceRegionsMultivariate1964}.
    For example, for $k=2$ and $\delta = 5.991$, the hypersphere contains \SI{95}{\%} of uncertainty realizations.
    
    After defining the latent-space hypersphere and mapping it to the data space, it is crucial to relate its size to the probability of feasibility.
    Because the normalizing flow is a diffeomorphic transformation, the probability mass within the hypersphere is preserved when mapped to the data space \citep{papamakarios2021normalizing}:
         \begin{equation*}
             \int_{\mathbf{l} \in \mathcal{L}} p_{l}(\mathbf{l}) d\mathbf{l} = \int_{\mathbf{y} \in \mathcal{Y}(\mathbf{c})} p_{y}(\mathbf{y}) d\mathbf{y}
         \end{equation*}
    In other words, the total probability mass contained within the hypersphere equals the total probability mass of the corresponding transformed set in the data space.
    This property allows us to interpret the size of the hypersphere in the latent space as a lower bound on the probability of feasibility in the data space.
    Since every point inside the admissible uncertainty set is guaranteed to be feasible, the probability of a random sample from the data distribution being feasible is at least as high as the probability of being in the admissible uncertainty set, and potentially higher if additional feasible points lie outside the admissible uncertainty set (see Figure \ref{fig:FlexInd}).

    An important limitation of normalizing flows to be mindful of is their preservation of topological properties \citep{papamakarios2021normalizing}.
    This means that if the data distribution contains multiple modes, but the base distribution only contains a single mode, the normalizing flow will assign non-zero probability to the empty space between modes in the data space, which introduces some error into the model.
    For our application of constructing admissible uncertainty sets, this means that if the data distribution contains holes, it cannot be perfectly approximated by a normalizing flow with a hole-free base distribution (cf. Section \ref{sec:TopologyPreservation}).

    \subsection{Normalizing flow architecture}
    We employ the RealNVP architecture \citep{dinh2017density}, which we have successfully applied to energy system time-series data in the past \citep{cramerprincipalcomponentdensity2022} to learn the transformation required to define the admissible uncertainty sets.
    \begin{figure}[H]
        \centering
        \includegraphics[width=0.5 \columnwidth]{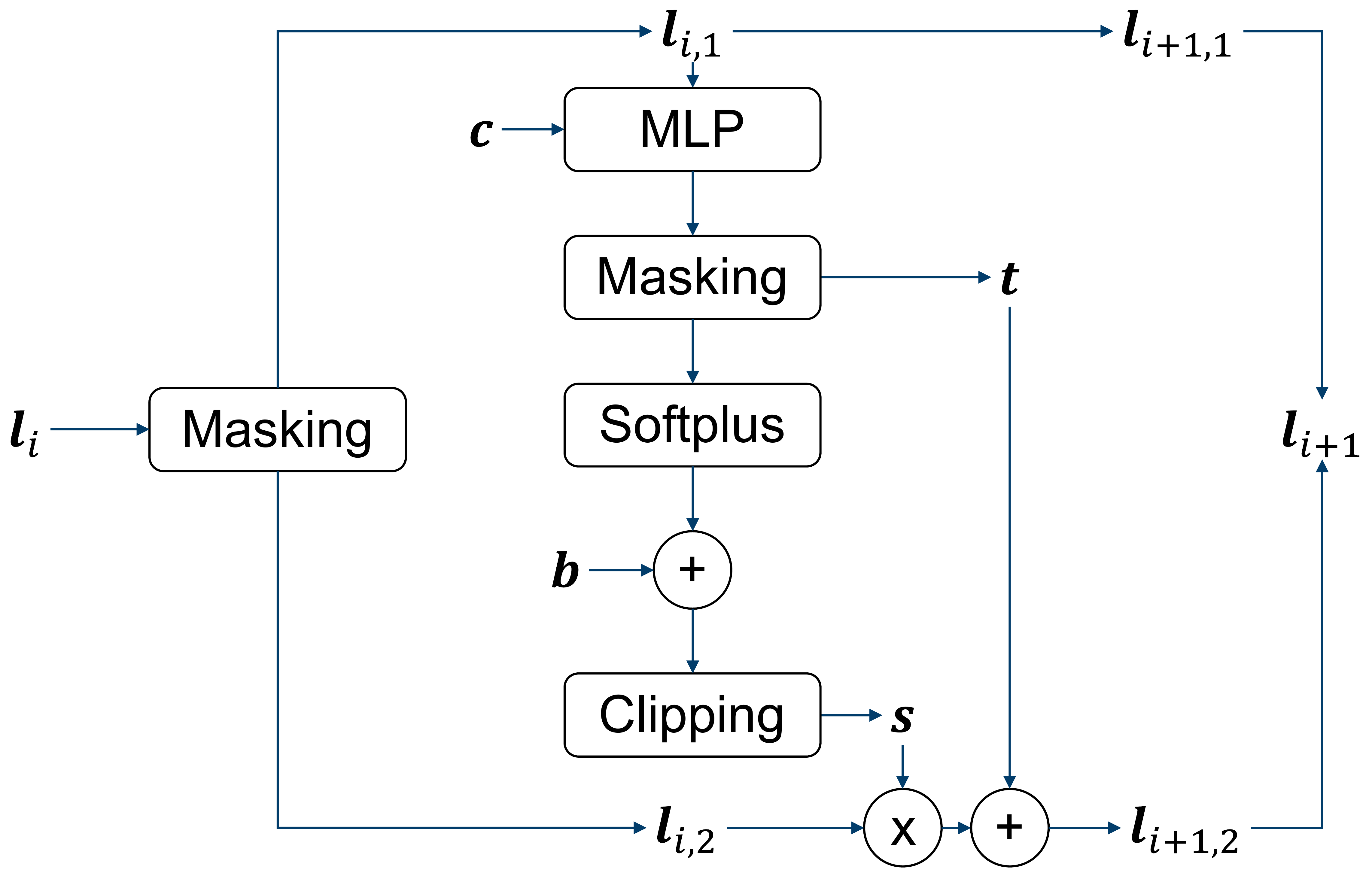}
        \caption{Illustration of a single RealNVP transformation block. The input $\mathbf{l}_{i}$ is split into two parts (masking): $\mathbf{l}_{i, 1}$ and $\mathbf{l}_{i, 2}$. The first part $\mathbf{l}_{i, 1}$ remains unchanged and is used to parameterize an affine transformation of $\mathbf{l}_{i, 2}$. The scale and translation parameters $\mathbf{s}$ and $\mathbf{t}$ are determined using an artificial neural network (ANN) conditioned on contextual information $\mathbf{c}$. The ANN comprises a multi-layer perception (MLP) whose outputs are split into the translation parameter $\mathbf{t}$ and a second component used to compute the scale $\mathbf{s}$. The scale vector is obtained by applying a softplus activation, adding a bias term $\mathbf{b}$, and clipping the result to the range $[0, 3]$. The transfomed output $\mathbf{l}_{i + 1, 2}$ is then combined with the untransformed part of the input $\mathbf{l}_{i + 1, 1}$ by inserting the entries of both subvectors into the original positions from $\mathbf{l}_{i}$, thereby forming the output of the RealNVP block $\mathbf{l}_{i + 1}$.}
        \label{fig:RealNVP}
    \end{figure}

    Figure \ref{fig:RealNVP} shows a RealNVP transformation block used in our implementation, based on the implementation by \cite{nflows} and \cite{torresFabricioArendTorresFlowConductor2025}.
    In each transformation block the input $\mathbf{l}_{i}$ is split into two parts (masking) $\mathbf{l}_{i, 1}$ and $\mathbf{l}_{i, 2}$.
    $\mathbf{l}_{i, 1}$ is left untransformed and used to parametrize an affine transformation of $\mathbf{l}_{i, 2}$.
    The transformation parameters scale $\mathbf{s}$ and translation $\mathbf{t}$ are determined by an ANN \citep{dinh2017density}, which takes contextual information $\mathbf{c}$ into account.
    The main block of the ANN determining the parameters is a multi-layer perceptron (MLP), whose size can be adjusted by changing the number of hidden units and the number of hidden layers.
    ReLU activations are used in the MLP, although other activation functions could be employed in principle.
    The MLP output is used in two parts: the translation parameter $\mathbf{t}$ is taken directly as one part of the output of the MLP, while the scale parameter $\mathbf{s}$ is determined by first applying the softplus activation function $Softplus(x) = \ln(1 + \exp{x})$ elementwise to the other part of the MLP output, adding a small bias $b = 0.001$, and clipping the values to the range $[0, 3]$.
    By alternating the part of the input that is left untransformed in each transformation block, highly expressive transformations can be learned \citep{dinh2017density}.

    To construct admissible uncertainty sets using normalizing flows, the trained model has to be embedded within the optimization problem.
    To facilitate model embedding, we modified the RealNVP implementation by replacing slicing operations with matrix multiplications.
    Specifically, the masking operations mentioned at the beginning of this section are implemented using matrix multiplication, e.g.,
    \begin{equation*}
    \mathbf{l}_{n, 1} = \mathbf{A}_{1} \mathbf{l}_{n},
    \end{equation*}
    where $\mathbf{A}_{1} \in \{0, 1\}^{|\mathbf{l}_{n, 1}| \times |\mathbf{l}_{n}|}$ contains a single one in each row, thereby selecting elements of $\mathbf{l}_{n}$ that belong to $\mathbf{l}_{n, 1}$.
    We then use the computational graph of the trained normalizing flow provided by ONNX \citep{bai2019} and embed each operation in the normalizing flow transformation as constraints into the optimization problem.
    A full-space formulation is used, exposing all the intermediate variables to the optimizer \citep{schweidtmannDeterministicGlobalOptimization2019}.
    The presence of clipping and ReLU activations \citep{2018arXiv180308375A} in the multi-layer perceptron leads to integer variables, while the softplus activation introduces nonlinear constraints.
    Consequently, the resulting problem is a mixed-integer nonlinear program that requires a solver capable of handling these challenging formulations.
    
\section{Illustrative examples}\label{sec:Illustrative_Example}
In this section, we first illustrate our approach using two examples based on well-known benchmark datasets for classification
\citep{scikit-learn}: the nonconvex ``Two Moons'' example and the ``Annulus'' example.
\subsection{Nonconvex data distributions}\label{sec:TwoMoons}
\begin{figure}[H]
    \centering
    \includegraphics[width=0.5\columnwidth]{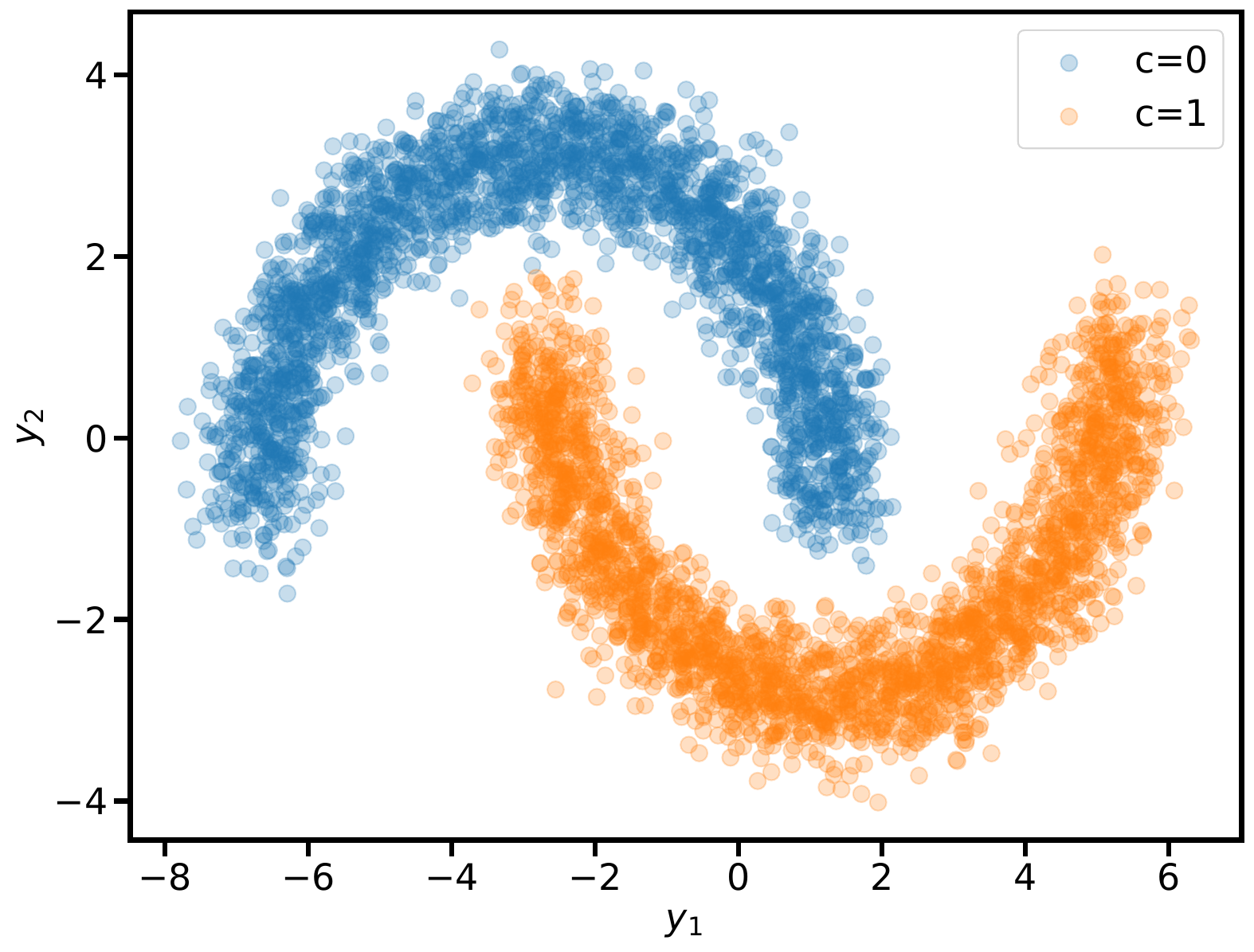}
    \caption{``Two-Moons'' \citep{scikit-learn} dataset for uncertain parameters, with a noise-level of $0.1$, scaled by a factor of $4$ and shifted by $\begin{bmatrix} -2.7, & -0.85 \end{bmatrix}^T$. The blue crescent moon is associated with the contextual information $c=0$ and the orange moon with $c=1$. $y_1$ and $y_2$ are the two dimensions of the uncertain variable vector $\mathbf{y}$.}
    \label{fig:TwoMoons}
\end{figure}

As data for the uncertain parameters, we use the two-dimensional ``Two-Moons'' dataset \citep{scikit-learn} shown in
Figure \ref{fig:TwoMoons}.
We choose a noise-level of $0.1$, scaled by a factor of $4$ and shifted by $\begin{bmatrix} -2.7, & -0.85 \end{bmatrix}^T$.
The dataset consists of two crescent moons, where one crescent moon is associated with the contextual information $c=0$ and the other with $c=1$.
This example with binary contextual information is chosen for illustrative purposes.
In Section \ref{sec:Case_Study}, the proposed method is applied using continuous contextual information, which is more typical in practical settings.
We train the normalizing flow model using a dataset of $100000$ samples and monitor the model performance during training using a validation set of $10000$ samples.
For additional information on the training settings, we refer to Section $1$ of the supplementary material.

\begin{figure}[H]
    \centering
    \includegraphics[width=0.5 \columnwidth]{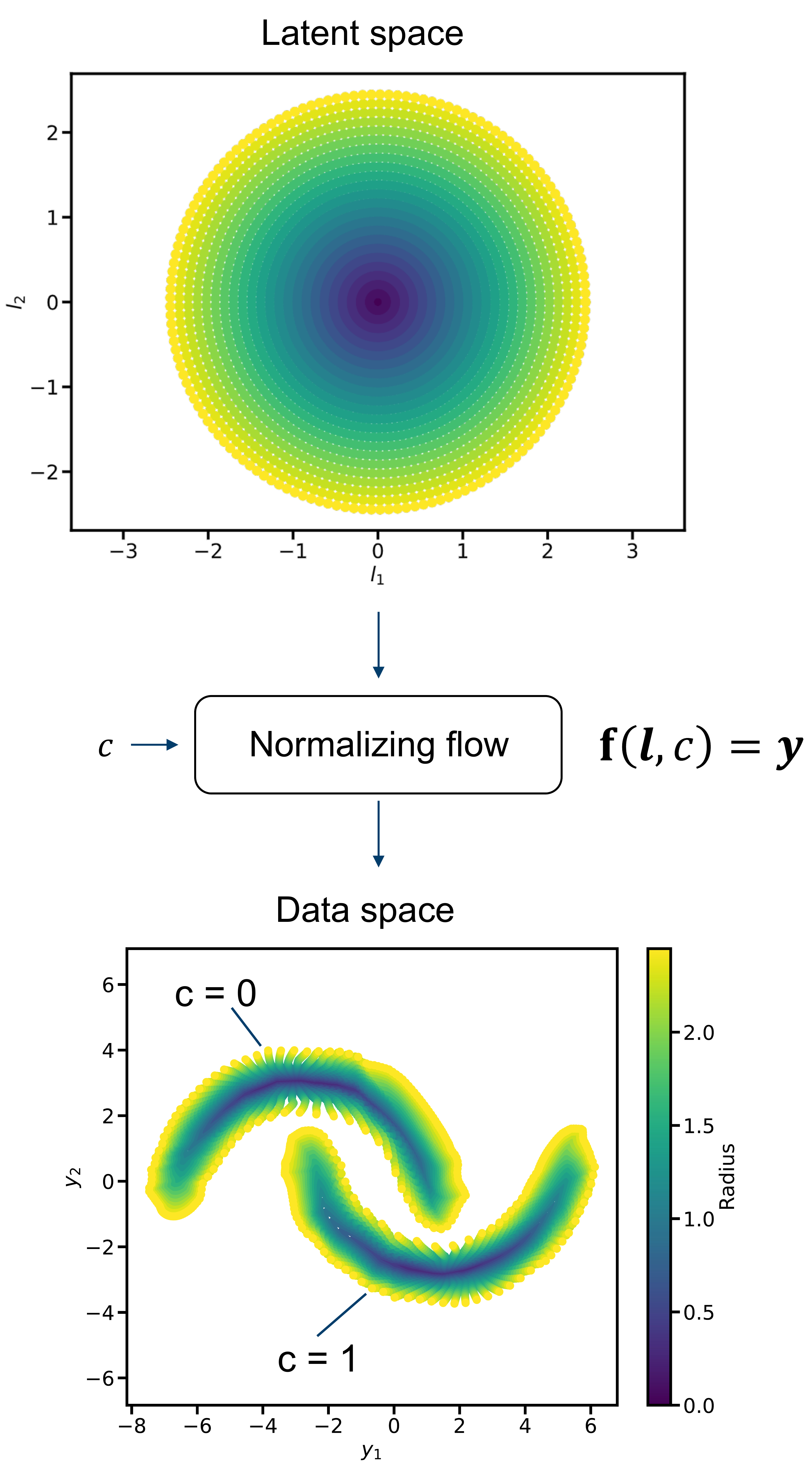}
    \caption{Illustration of the transformation of the admissible uncertainty set from the latent space to the data space. Color indicates the radius of the points in the latent space.}
    \label{fig:Transformation}
\end{figure}

Figure \ref{fig:Transformation} illustrates our approach.
The two-dimensional hypersphere admissible uncertainty set in the latent space is transformed by a conditional normalizing flow to the data space.
Depending on the value of the contextual information $c$, this leads to two crescent moon-shaped admissible uncertainty sets in the data space.
The color of the points illustrates the distance of the points in the latent space from the origin.
Points on the edge of the admissible uncertainty set in the latent space also end up at the edge of the admissible uncertainty set in the data space.

We use a scaled and shifted version of the Himmelblau function \citep{himmelblau1972applied}, i.e.,
\begin{equation*}
    \begin{split}
            h(\mathbf{y}) = \left(0.53\left(y_{1} + 0.9\right)^2 + y_{2} - 11\right)^2 + \left(0.53\left(y_{1} + 0.9\right) + \left(y_{2}^2 - 7\right)\right)^2 ,
    \end{split}
\end{equation*}
to define the constraint
\begin{equation*}
    g(\mathbf{y}) = 10 - h(\mathbf{y}) \le 0
\end{equation*}
of the embedded maxmin problem in the flexibility index problem (Problem \eqref{cond_flex_ind}).
For simplicity, we do not introduce decision variables $\mathbf{x}$ or recourse variables $\mathbf{z}$ in our illustrative example; instead, we focus solely on determining the CFI $\delta_{cond}^{*}$:
\begin{align*}
            \delta_{cond}^{*}(c) = &\underset{\delta \in \mathbb{R}}\max \quad \delta \\
            &\text{s.t.} \quad g(\mathbf{y}) \le 0 \ \forall \mathbf{y} \in \mathcal{Y}_{cond}(c, \delta)
\end{align*}
The semi-infinite constraint can be formulated as the optimization problem \citep{blankenshipinfinitelyconstrainedoptimization1976}
\begin{align*}
            & \underset{\mathbf{y} \in \mathcal{Y}_{cond}(c, \delta)}\max \quad g(\mathbf{y}) \le 0, 
\end{align*}
with the feasible set $\mathcal{Y}_{cond}(c, \delta) = \left\{ \mathbf{y} \in \mathbb{R}^{2} | \mathbf{l} \in \mathbb{R}^{2}, \left\lVert \mathbf{l} \right\rVert_{2}^2 - \delta \le 0, \mathbf{y} = \mathbf{f}(\mathbf{l}, c) \right\}$.
Since the feasible set is defined by coupling constraints, i.e., constraints that depend on the upper-level variable $\delta$, we have a generalized semi-infinite program (GSIP).
To solve the GSIP, we form the SIP relaxation by moving the coupling inequality constraints into the objective function.
Note, that we make the assumption that the SIP relaxation has the same objective value as the original GSIP \citep{mitsosglobaloptimizationgeneralized2015} and reads
\begin{align*}
            \underset{\mathbf{l} \in \mathbb{R}^{2}, \mathbf{y} \in \mathcal{Y}_{SIP}(\mathbf{l}, c)}\max & \min\left\{g\left(\mathbf{y}\right), \delta - \left\lVert \mathbf{l} \right\rVert_{2}^2 \right\} \le 0,
\end{align*}
with the feasible set $\mathcal{Y}_{SIP}(\mathbf{l}, c) = \left\{\mathbf{y} \in \mathbb{R}^{2} | \mathbf{y} = \mathbf{f}(\mathbf{l}, c) \right\}$.

Finally, we end up with the CFI problem
\begin{equation}
    \begin{aligned}\label{eq:CFI}
                \delta_{cond}^{*}(c) = 
                \underset{\delta \in \mathbb{R}}\max & \quad \delta \\
                \text{s.t.} & \quad              \underset{\mathbf{l} \in \mathbb{R}^{2}, \mathbf{y} \in \mathcal{Y}_{SIP}(\mathbf{l}, c)}\max \min\left\{g\left(\mathbf{y}\right), \delta - \left\lVert \mathbf{l} \right\rVert_{2}^2\right\} \le 0,
    \end{aligned}
\end{equation}
which we solve using the adaptive discretization algorithm by \cite{blankenshipinfinitelyconstrainedoptimization1976}, where we use a feasibility tolerance of $0.05$ for the semi-infinite constraint, i.e., Problem \eqref{eq:CFI} is considered feasible if the lower-level problem satisfies $\underset{\mathbf{l} \in \mathbb{R}^{2}, \mathbf{y} \in \mathcal{Y}_{SIP}(\mathbf{l}, c)}\max \min\left\{g\left(\mathbf{y}\right), \delta - \left\lVert \mathbf{l} \right\rVert_{2}^2\right\} \le 0.05$.

\begin{figure}[H]
    \centering
    \includegraphics[width=0.5\columnwidth]{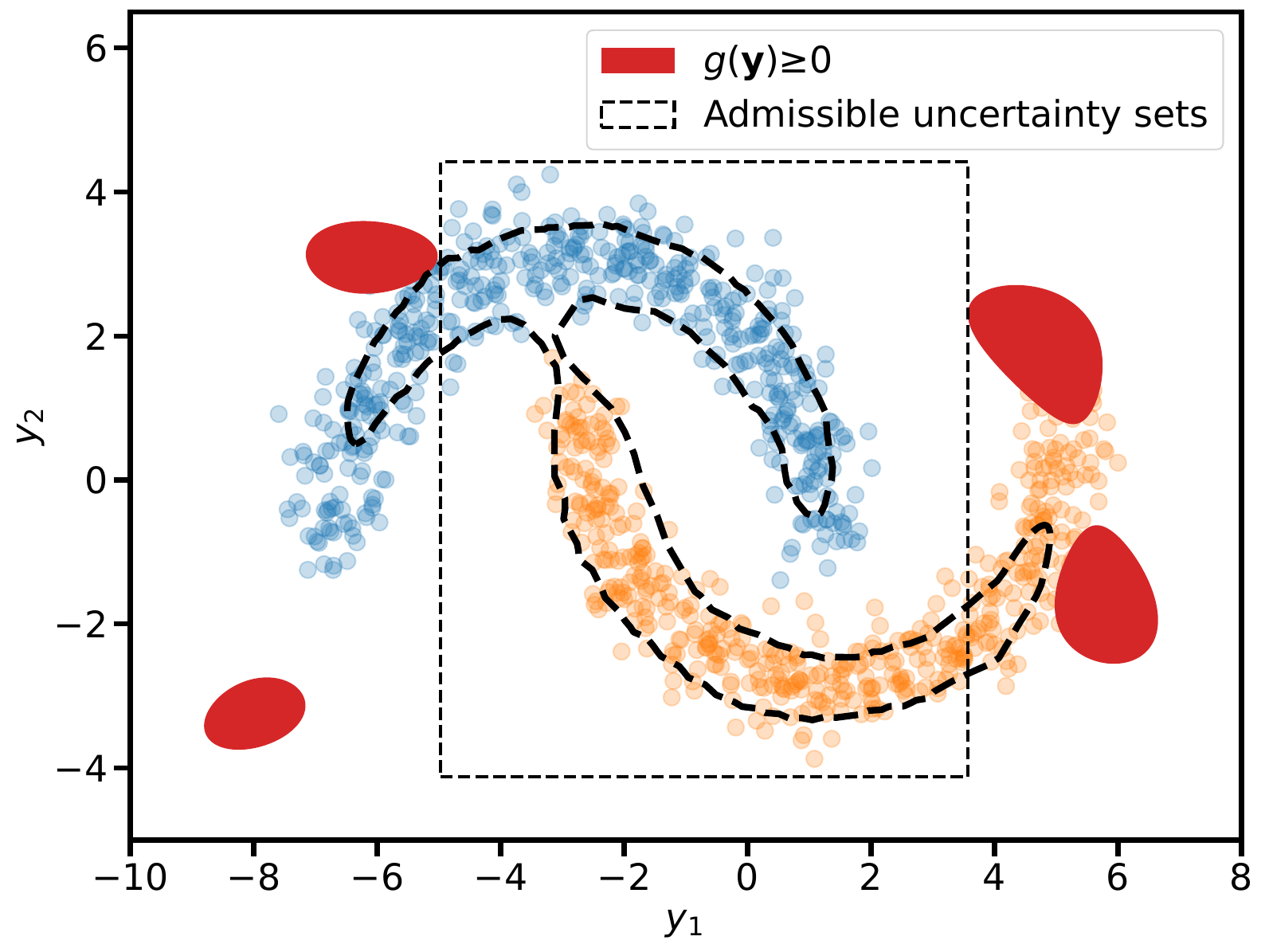}
    \caption{Unconditional normalizing flow-based and hypercube admissible uncertainty sets plotted in the data space. Colored points indicate samples from the data distribution. Blue color corresponds to $c=0$ and orange color corresponds to $c=1$. Red areas mark infeasible regions.}
    \label{fig:TwoMoonsResUncon}
\end{figure}

To demonstrate our method, we train an unconditional and a conditional normalizing flow on the ``Two-Moons'' dataset, each with $5$ RealNVP blocks, containing an MLP with $1$ hidden layer of $12$ hidden units, and use them to implement the transformation  $\mathbf{f}(\mathbf{l}, c)$ in Problem \eqref{eq:CFI}, with the transformation of the unconditional flow reading $\mathbf{f}(\mathbf{l})$.
For comparison, we determine hypercube admissible uncertainty sets in the \emph{data space}, centered at the unconditional and conditional means of the data distribution $\mathbf{y}_{cube, 0}(c)$ by modifying Problem \eqref{cond_flex_ind}:
\begin{align*}
            \delta_{cond}^{*}(c) = 
            \underset{\delta \in \mathbb{R}}\max & \quad \delta \\
            \text{s.t.} & \quad              \underset{\mathbf{y}_{cube} \in \mathbb{R}^{2}}\max \min\left\{g\left(\mathbf{y}_{cube}\right), \delta - \left\lVert \mathbf{y}_{cube}  - \mathbf{y}_{cube, 0}(c)\right\rVert_{\infty}\right\} \le 0
\end{align*}
First, we demonstrate the unconditional flexibility index.
Figure \ref{fig:TwoMoonsResUncon} shows the resulting admissible uncertainty sets, together with samples from the data distribution and the infeasible regions.
The normalizing flow-based admissible uncertainty set somewhat mimics the shape of the data distribution, whereas the hypercube admissible uncertainty set contains areas where no historical data is present.
To enable a quantitative assessment, we estimate the coverage of the admissible uncertainty sets by drawing random samples from the data distribution and calculating the proportion that falls within the admissible uncertainty sets.
The normalizing flow admissible uncertainty set achieves a coverage of \SI{61}{\%} while the hypercube achieves a coverage of \SI{69}{\%}.
In general, the achieved coverage depends strongly on the specific problem, as the interplay between the shape of the data distribution and the admissible uncertainty region determines the maximum attainable admissible uncertainty set size, and therefore the flexibility index.
By modifying the admissible uncertainty region, for example, to a small circle around the center of the hypercube or around the center of the normalizing flow admissible uncertainty set, instances can be constructed in which either the data-driven admissible uncertainty set or the hypercube outperforms the other.
This is an important consequence of the CFI approach, which learns the center of the admissible uncertainty set, whereas in the traditional flexibility index, the center (i.e., the nominal parameter value) is chosen by the user according to some criterion, such as the expected value of the parameter realization.
Consequently, no universal statement can be made regarding the superiority of data-driven admissible uncertainty sets over simple admissible uncertainty sets.
Nevertheless, the data-driven admissible uncertainty sets ensure that constraints that lie in regions where no historical data is present do not limit the flexibility index.

\begin{figure}[H]
    \centering
    \includegraphics[width=0.5\columnwidth]{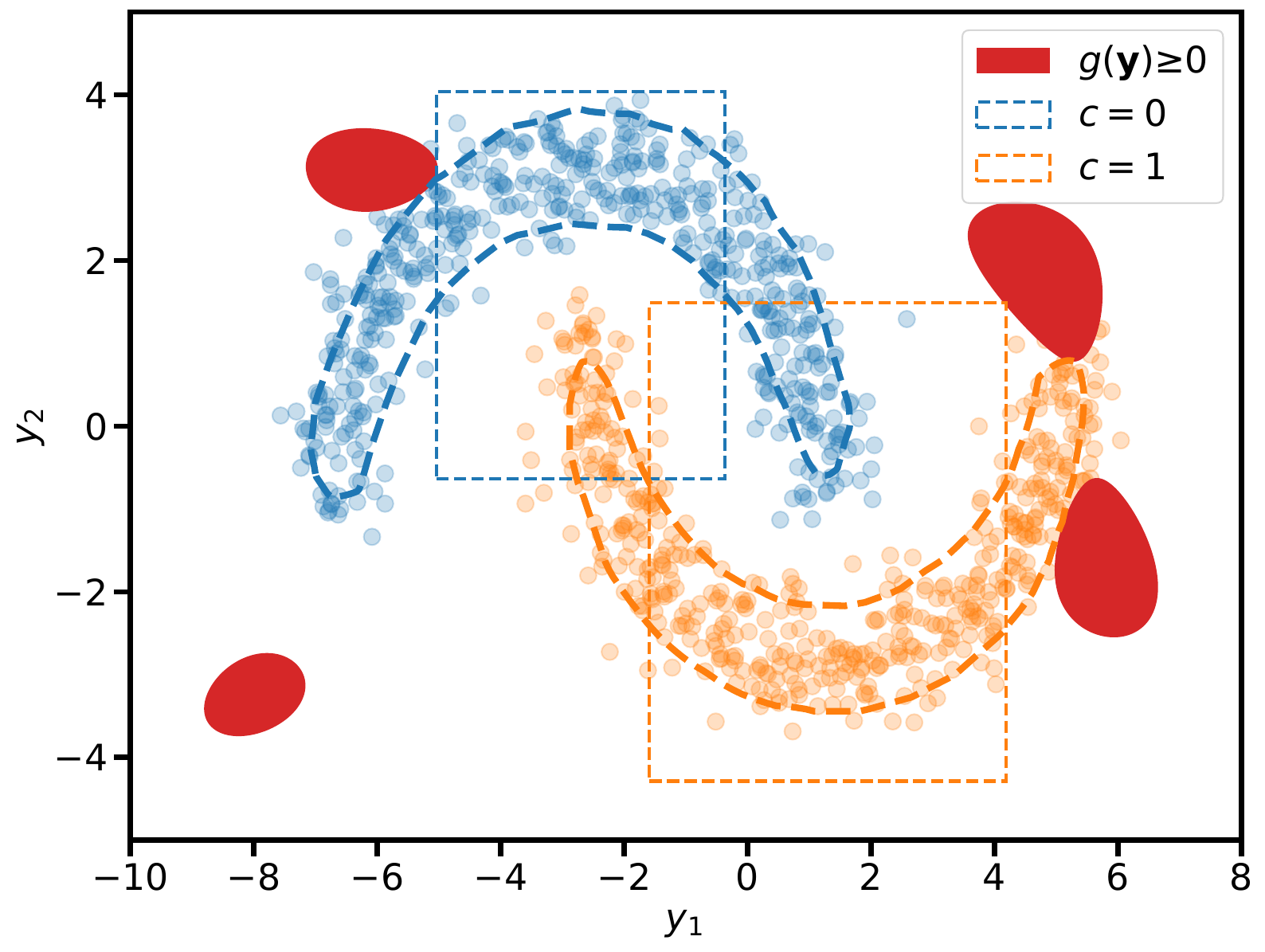}
    \caption{Conditional normalizing flow-based and hypercube admissible uncertainty sets plotted in the data space. Colored points indicate samples from the data distribution. Blue color corresponds to $c=0$ and orange color corresponds to $c=1$. Red areas mark infeasible regions.}
    \label{fig:TwoMoonsRes}
\end{figure}

Next, we demonstrate the conditional flexibility index.
Figure \ref{fig:TwoMoonsRes} shows the resulting admissible uncertainty sets, together with samples from the data distribution and the infeasible regions.
Note that for both the normalizing flow and the hypercube approaches, the conditional admissible uncertainty sets include regions not present in the corresponding unconditional sets (cf. Figure \ref{fig:TwoMoonsResUncon}), and vice versa.
This indicates, that no general statement can be made about one approach consistently outperforming the other.
However, similar to the comparison between simple and data-driven sets, conditional admissible uncertainty sets have the advantage of excluding regions where no uncertainty realizations exist for a given contextual information, ensuring that the flexibility index is not limited by infeasibilities in such regions.

Comparing the two conditional approaches, normalizing flow–based admissible uncertainty sets achieve higher conditional coverage of \SI{79}{\%} and \SI{75}{\%} compared to \SI{40}{\%} and \SI{52}{\%} for the hypercube admissible uncertainty sets.
This conditional coverage estimation requires access to the original data distribution for sampling, which is rarely available in a practical setting.
As mentioned in Section \ref{sec:NFUC}, the conditional coverage of the normalizing flow-based admissible uncertainty sets, however, can be estimated from the calculated flexibility index.
With the flexibility indices calculated by solving Problem \eqref{eq:CFI}, $\delta_0 = 3.12$ and $\delta_1 = 2.84$, we obtain analytical coverages of \SI{79}{\%} and \SI{76}{\%}, respectively, agreeing with the empirical estimates.

The conditional coverage probability provides a lower bound on the conditional probability of feasibility, provided that the underlying data distribution is accurately approximated.
However, if the normalizing flow fails to represent the true distribution adequately, the conditional coverage probability may deviate from the true conditional probability of feasibility.
Ensuring sufficient representational capacity of the normalizing flow is therefore essential.

The quality of the distributional approximation can be assessed in a problem-specific manner.
For low-dimensional datasets, visual inspection of the learned density may be informative, whereas for higher-dimensional data, comparisons of marginal probability density functions may be used \citep{cramerValidationMethodsEnergy2022}.
In addition, the log-likelihood computed on the validation set can be used to compare models with different representational capacities \citep{kingma2018glow}: If the validation log-likelihood increases as model capacity increases, this may indicate that models with lower capacity are underfitting.

\begin{figure}[H]
    \centering
    \includegraphics[width=0.5\columnwidth]{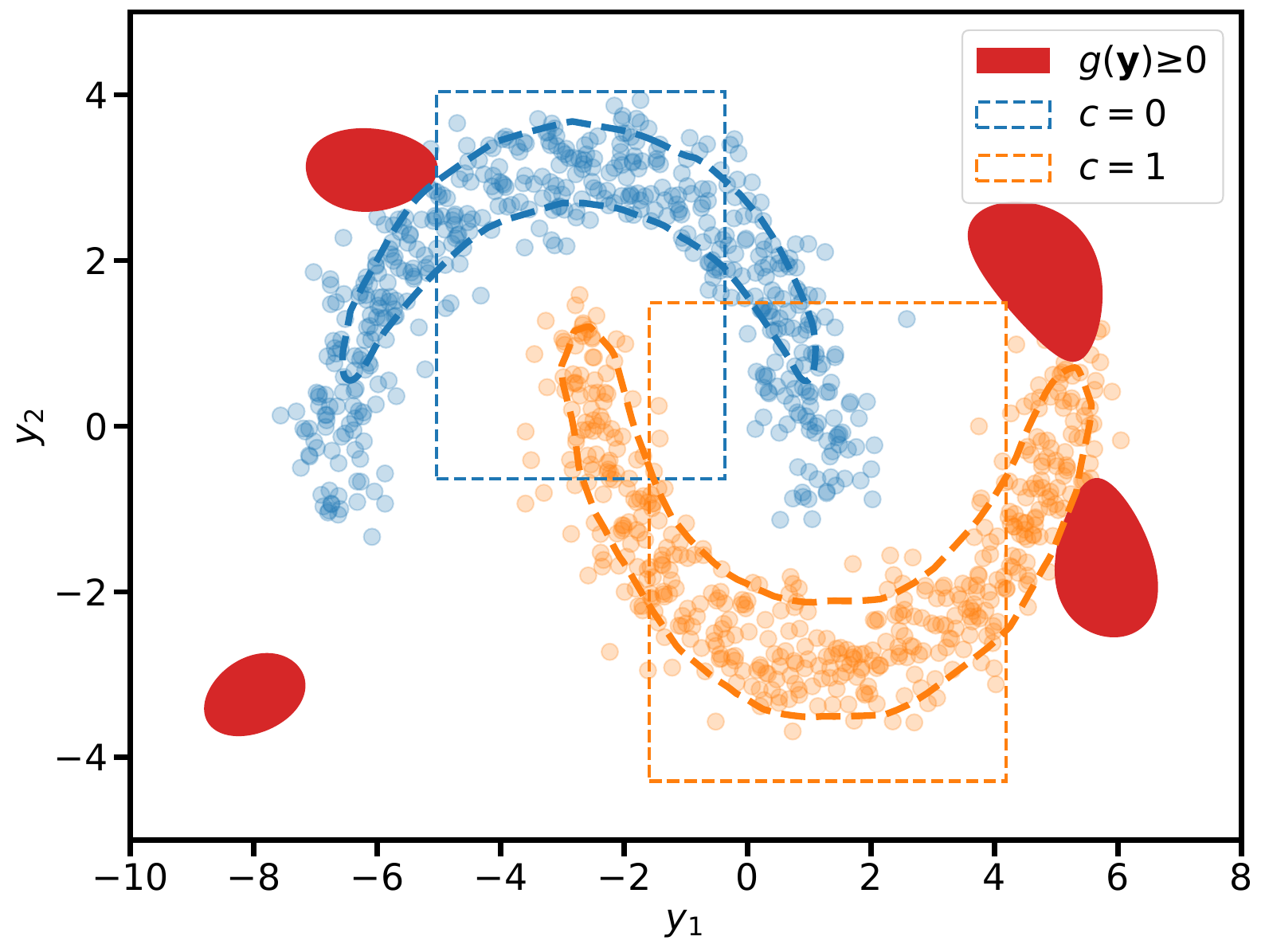}
    \caption{Normalizing flow-based and hypercube admissible uncertainty sets plotted in the data space. Colored points indicate samples from the data distribution. Blue color corresponds to $c=0$ and orange color corresponds to $c=1$. Red areas mark infeasible regions. The normalizing flow is trained with a different random seed than the normalizing flow used to generate Figure \ref{fig:TwoMoonsRes}. The admissible uncertainty region for $c=0$ is smaller than in Figure \ref{fig:TwoMoonsRes} due to different normalizing flow parameters, illustrating the variability inherent in the CFI approach.}
    \label{fig:TwoMoonsRes_2}
\end{figure}

The shape of the admissible uncertainty sets and hence the CFI depend on the trained normalizing flow model, and there is some variability when training a model of the same capacity with different random seeds.
Figure \ref{fig:TwoMoonsRes_2} shows the resulting admissible uncertainty sets for a normalizing flow with the same capacity but a different random seed.
The admissible uncertainty set for $c=0$ is significantly smaller, caused by slightly different parameters.
However, the conditional coverage of \SI{51}{\%} still surpasses the coverage of the hypercube admissible uncertainty set of \SI{40}{\%}.

\begin{figure}[H]
    \centering
    \includegraphics[width=0.9 \columnwidth]{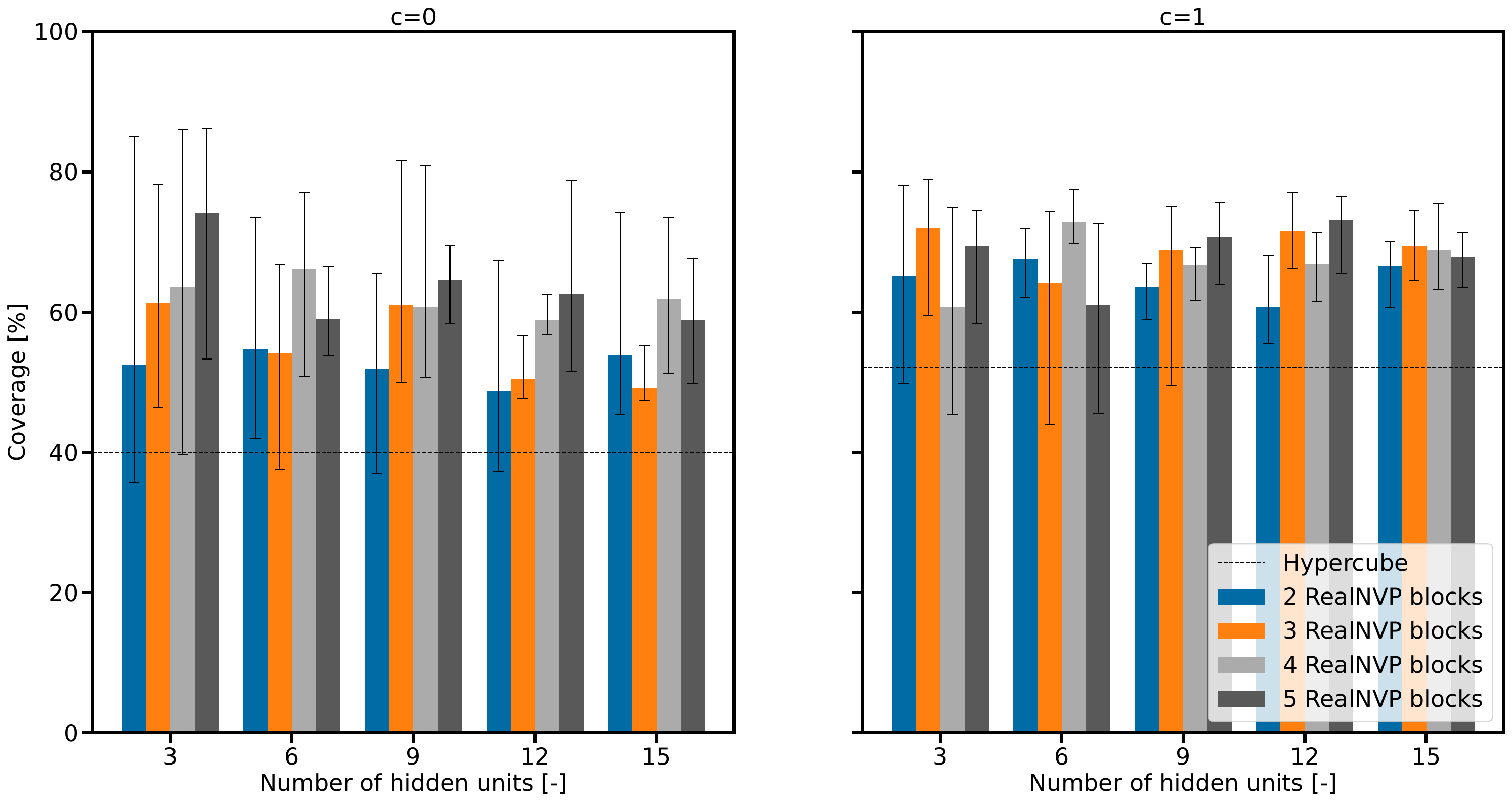}
    \caption{Average conditional coverage probability achieved by normalizing flow-based admissible uncertainty sets for varying number of RealNVP blocks and hidden units.
    The average is determined over $5$ runs with different random seeds. Whiskers indicate the highest and lowest coverage probability achieved. Hypercube admissible uncertainty set coverage is shown as a comparison (dashed line).}
    \label{fig:Variability}
\end{figure}

Figure \ref{fig:Variability} shows the average conditional coverage probability achieved by the normalizing flow-based admissible uncertainty sets for $c=0$ (top) and $c=1$ (bottom).
The average is calculated over $5$ runs with different random seeds, whiskers indicate the highest and lowest coverage achieved.
The number of RealNVP blocks and hidden units in the MLPs, which parametrize the transformation parameters $\mathbf{s}$ and $\mathbf{t}$, is varied.
The variability between runs is higher for $c=0$ than for $c=1$, indicated by the bigger span between the best and worst runs.
The high variability in conditional coverage, combined with a lower average coverage, indicates that the representational capacity of the normalizing flow is insufficient when only a small number of hidden units or RealNVP blocks are used.
The number of RealNVP blocks has a stronger influence on average performance, especially for $c=0$, while the number of hidden units shows no consistent trend, though variability decreases as the number of hidden units increases.
Once a sufficient capacity is achieved, i.e., $6$ or more hidden units and $4$ or more RealNVP blocks, performance differences appear to be random.

Generally, the difference in conditional coverage between the normalizing flow and the hypercube admissible uncertainty set depends strongly on the shape of the underlying data distribution and the constraints.
In the ``Two-Moons'' example, the normalizing flow-based admissible uncertainty set achieves higher average conditional coverage than the hypercube-based admissible uncertainty set.
However, the coverage varies with the random seed, and there are a few instances where the normalizing flow-based conditional coverage is actually lower than the hypercube coverage.
Nevertheless, each admissible uncertainty set is rigorous in the sense that feasibility is guaranteed.
A higher conditional coverage can be achieved by training multiple normalizing flows and choosing the flow that achieves the highest coverage.

\begin{figure}[H]
    \centering
    \includegraphics[width=0.5 \columnwidth]{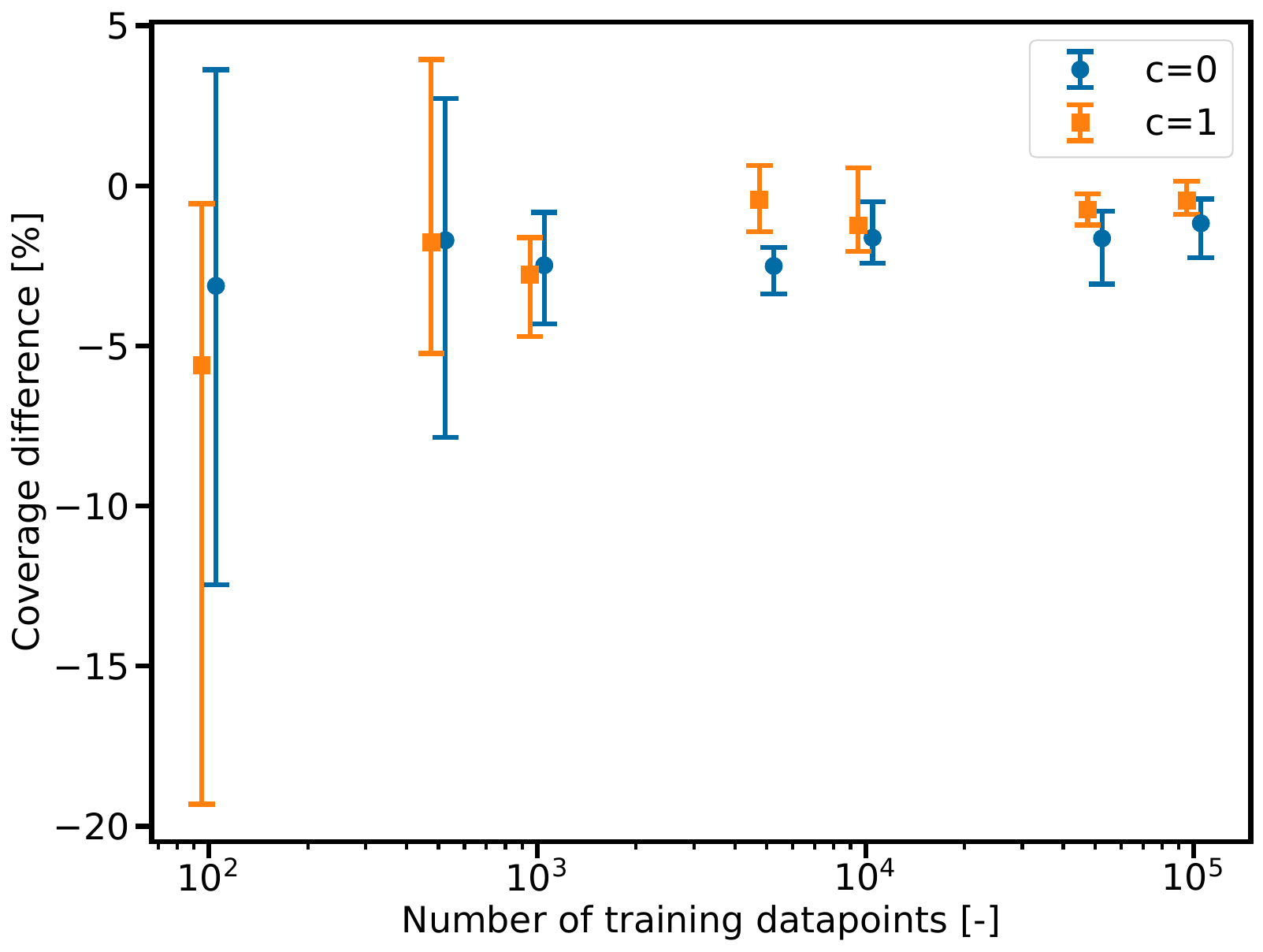}
    \caption{Coverage difference, i.e., difference between analytical and empirical conditional coverage for different training dataset sizes.
    Negative values indicate that the analytical conditional coverage is smaller than the empirical conditional coverage, i.e., the true conditional coverage is likely underestimated.
    The average is determined over $5$ runs with different random seeds. Whiskers indicate the highest and lowest coverage difference achieved.}
    \label{fig:estimation_acc}
\end{figure}

In the aforementioned experiments, the normalizing flow models were trained on a large dataset consisting of $100,000$ samples to isolate the performance of the proposed method from the effects of limited training data.
This, however, raises the question of how much data is required to obtain an accurate model.
To this end, normalizing flow models with $3$ RealNVP blocks and $12$ hidden units were trained on datasets of varying sizes.
Figure \ref{fig:estimation_acc} shows the difference between analytical and empirical conditional coverage, i.e., the coverage difference, for the respective dataset sizes.
A negative value indicates that the analytical conditional coverage is smaller than the empirical conditional coverage, which indicates that the analytical coverage likely underestimates the true conditional coverage.
For small datasets, the coverage difference exhibits a larger magnitude and variability.
Both magnitude and variability decrease as the number of training samples increases.
With $1,000$ samples, the coverage difference is already small, ranging between \SI{0}{\%} and \SI{-5}{\%}.
The small coverage discrepancy suggests that the normalizing flow model provides an accurate approximation of the underlying data distribution at this sample size.

Finally, we analyze the effect of normalizing flow model size on computational performance.
To this end, we measure the wall-clock time required to solve Problem \eqref{eq:CFI} for $c=0$ and $c=1$ using five different random seeds.
The average wall-clock time is then computed over ten runs in total using a desktop computer running Microsoft Windows 10 Enterprise version $10.0.17763$ with a $4$-core/$4$-thread Intel i5-4570 CPU with \SI{3.2}{\giga\hertz}/\SI{3.6}{\giga\hertz} base/turbo frequency and \SI{16}{\giga\byte} of RAM.
Furthermore, we use Gurobi \citep{gurobi} version $12.0.3$ to solve all optimization subproblems in the adaptive discretization-based algorithm \citep{blankenshipinfinitelyconstrainedoptimization1976}.
Different parameter settings were used in our experiments; however, deviations from the default configuration do not seem to significantly affect the performance.
In initial experimentation, in some instances, the solver reported suboptimal solutions as optimal. According to the Gurobi documentation \citep{gurobi} this may be an indicator of numerical issues.
This behavior was mitigated by setting the parameter `NumericFocus' to $2$, 
which, according to the Gurobi documentation \citep{gurobi} increases the time spent in checking the numerical accuracy of intermediate results and the effort spent to avoid numerical issues at the cost of speed.
After adjusting `NumericFocus', we did not encounter erroneous results again.

Solving Problem \eqref{eq:CFI} for the ``Two Moons'' example takes between $5$ and $9$ iterations of the adaptive discretization algorithm, with an average of $6.8$ iterations.
No trend was observed for the number of iterations with respect to the number of hidden units or RealNVP blocks.
For each instance, the runtimes reported by Gurobi for the upper- and lower-level subproblems were recorded and aggregated.
The lower-level problems, i.e., the embedded maxmin problem in Problem \eqref{eq:CFI} containing the normalizing flow, account for between \SI{94.9}{\%} and \SI{100}{\%} of the total runtime, i.e., the sum of the lower-level and upper-level runtimes.
This indicates that the embedded maxmin problem is the dominant contributor to overall wall-clock time.

\begin{figure}[H]
    \centering
    \includegraphics[width=0.5\columnwidth]{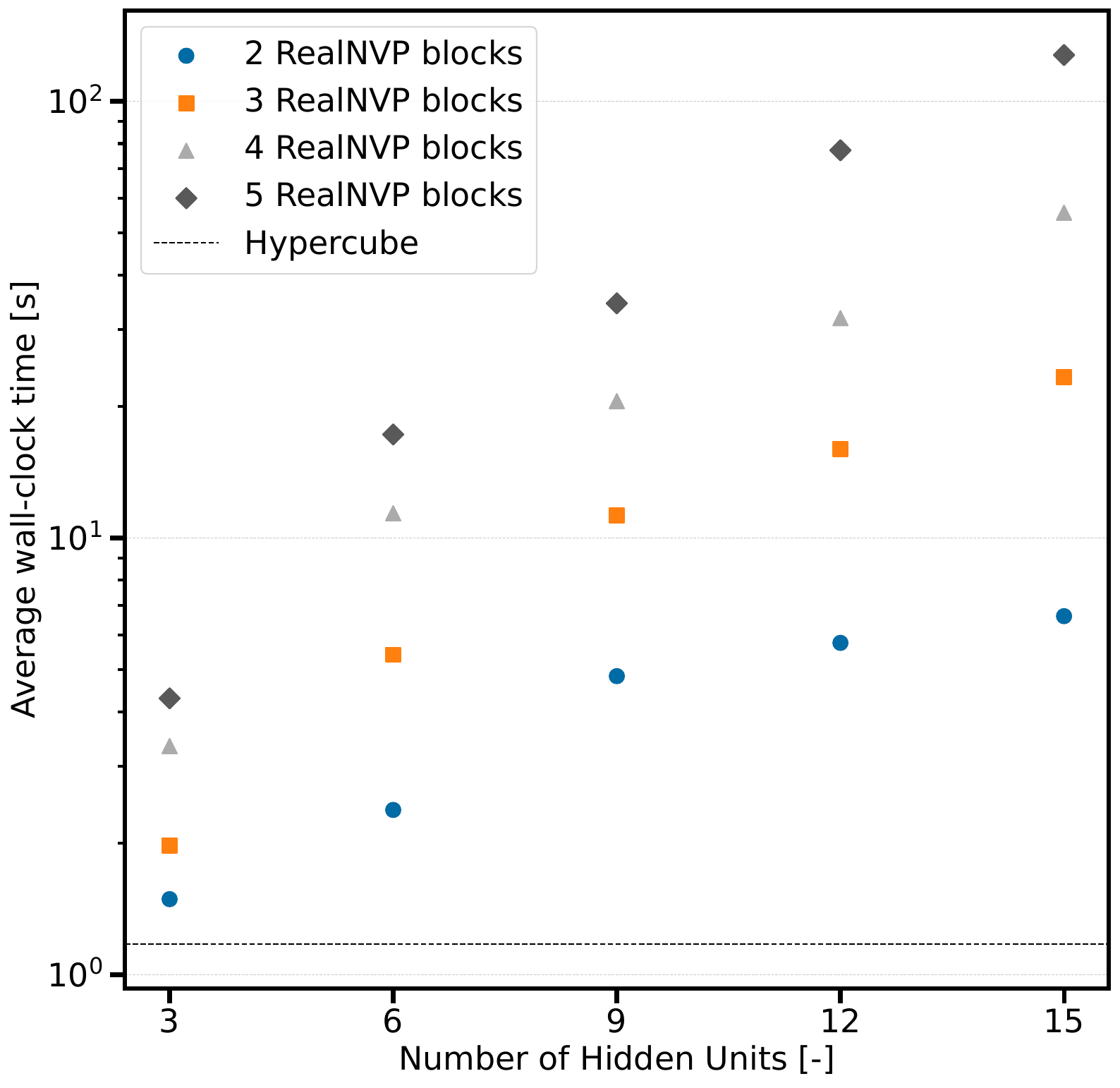}
    \caption{Average wall-clock time to solve Problem \eqref{eq:CFI} for varying number of RealNVP blocks and hidden units.
    The average is computed over ten runs, i.e., $c=0$ and $c=1$ for five different random seeds.
    The average wall-clock time to solve the hypercube-based approach is shown as a comparison (dashed line).}
    \label{fig:Runtime}
\end{figure}

Figure \ref{fig:Runtime} shows the resulting average wall-clock time for varying numbers of RealNVP blocks and hidden units.
A clear trend emerges: the average wall-clock time increases exponentially with the number of hidden units.
A similar trend occurs for the number of RealNVP blocks (not shown in the figure).
The observed increase in wall-clock time results from the growing number of constraints and variables introduced into the optimization problem as the normalizing flow model becomes larger.
In contrast to standard feedforward networks, the RealNVP architecture introduces an additional source of nonlinearity beyond the activation function, namely the softplus operation used to compute the scaling factor $\mathbf{s}$ (cf. Figure~\ref{fig:RealNVP}).
The runtimes for the lower-level problem reported by Gurobi \citep{gurobi} range from \SI{0.028}{\second} to \SI{47.685}{\second}.
We refer to Section $2$ in the supplementary material for detailed information on the introduced number of constraints and variables, as well as average lower-level runtimes.

The hypercube-based approach is included in Figure \ref{fig:Runtime} for comparison and exhibits lower average wall-clock times than all considered normalizing flow configurations.
Only problems with small embedded normalizing flow models exhibit wall-clock times comparable to those of the hypercube-based approach.
In this illustrative example, the embedded maxmin problem is structurally simple, consisting of only a single quadratic constraint aside from the admissible uncertainty set representation.
Consequently, the relative increase in wall-clock time due to embedding the normalizing flow may be more pronounced here than for maxmin problems with a more complex structure.
In summary, the average time required to solve Problem \eqref{eq:CFI} with a normalizing flow-based admissible uncertainty set grows exponentially with the size of the normalizing flow model and is generally larger than if a hypercube admissible uncertainty set is used.

\subsection{Topological property preservation}\label{sec:TopologyPreservation}
To illustrate the topological property preservation described in Section \ref{sec:NFUC}, we introduce an example of an annulus-shaped admissible uncertainty set.
Here, we limit ourselves to a single unconditional data set.
Specifically, we utilize the larger annulus from the ``circles'' dataset, generated using scikit-learn \citep{scikit-learn} with a noise level of $0.1$.
This dataset contains a hole, and, therefore, it cannot be approximated perfectly by a normalizing flow with a Gaussian base distribution.

As a constraint, we impose that the uncertainty realization needs to lie outside a circle with radius $r = 0.5$.
As in the example in Section \ref{sec:TwoMoons}, we do not have decision and recourse variables $\mathbf{x}$ and $\mathbf{z}$, and solve the problem in an identical way.
Furthermore, we also use the same hyperparameters to train the unconditional normalizing flow model.

\begin{figure}[H]
    \centering
    \includegraphics[width=0.5\columnwidth]{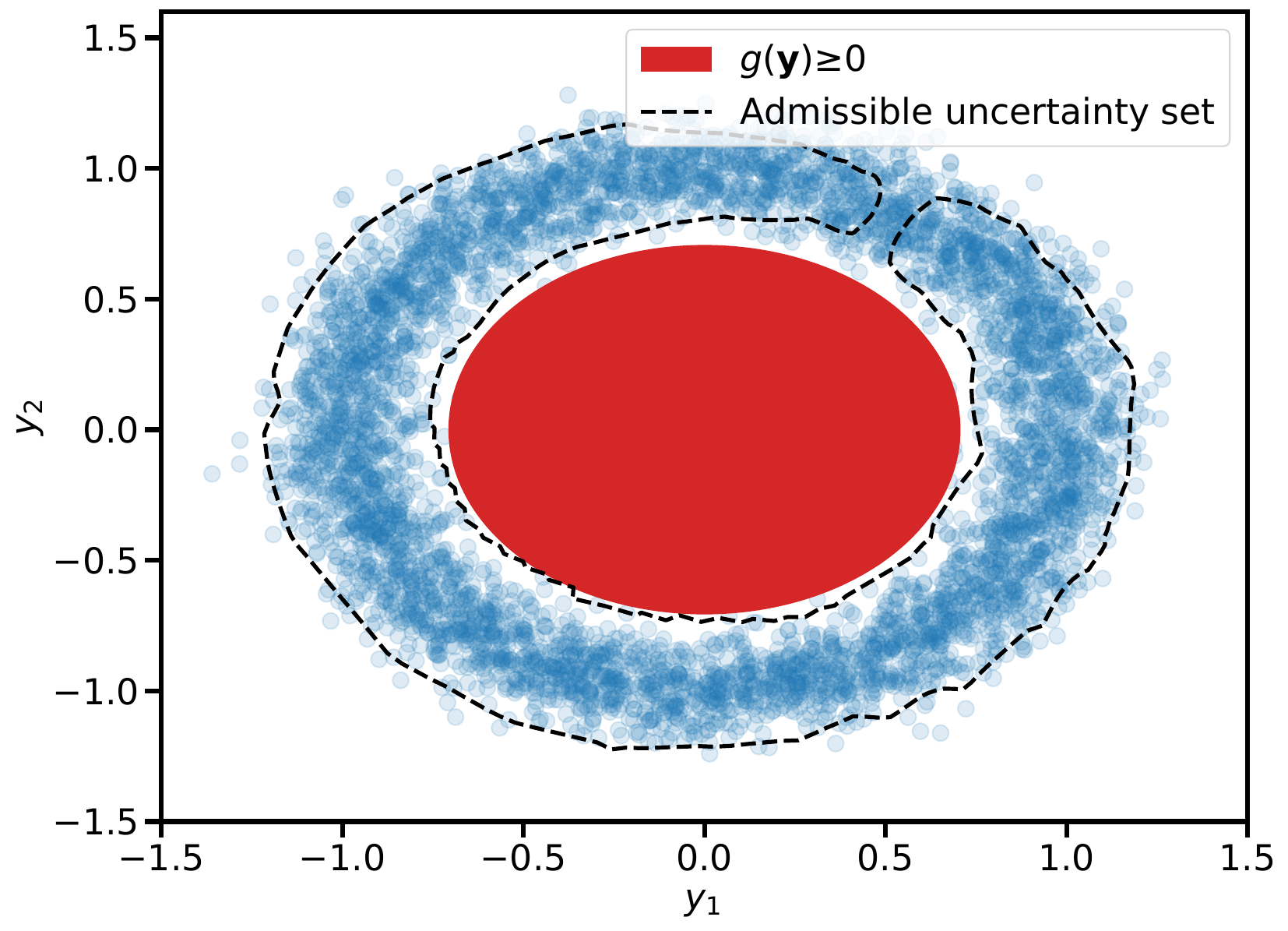}
    \caption{Normalizing flow-based admissible uncertainty set and samples from the data distribution (blue circles). Infeasible areas are marked red. The normalizing flow preserves topological properties; hence, the annulus is approximated by a C-shaped admissible uncertainty set, i.e., an annulus which is missing a small angular section.}
    \label{fig:annulus}
\end{figure}

Figure \ref{fig:annulus} illustrates the normalizing flow-based admissible uncertainty set, obtained by solving the flexibility index problem.
Since the Gaussian base distribution of the normalizing flow does not contain a hole, the admissible uncertainty set is approximated as a C-shaped admissible uncertainty set.
Although the normalizing flow-based admissible uncertainty set cannot perfectly approximate the data distribution, it still achieves a high coverage of \SI{79}{\%}.

If the underlying data distribution consists of disconnected clusters, the normalizing flow–based admissible uncertainty set remains a single connected set that bridges the clusters.
This behavior is illustrated in Figure \ref{fig:TwoMoonsResUncon}, where the two disconnected crescent-shaped regions are approximated by a single admissible uncertainty set containing a narrow connecting region.
If this connecting region intersects infeasible areas, the resulting admissible uncertainty set may lead to a significant underestimation of the probability of feasibility.

Disconnected datasets can be addressed by clustering the historical data and constructing a separate admissible uncertainty set for each cluster.
However, such an approach requires solving multiple embedded problems, i.e., one per cluster, and complicates the interpretation of the resulting conditional flexibility index since each cluster is associated with a different latent-space hypersphere radius.
The overall conditional coverage probability can still be estimated as the weighted sum of the cluster-specific conditional coverage probabilities, where the weights correspond to the probability mass of each cluster.
\section{Security-constrained unit commitment}\label{sec:Case_Study}
In the following section, we apply the CFI to a security-constrained unit commitment (SCUC) problem.
Several studies have demonstrated successful applications of the flexibility index in power systems, where it serves as a tool to ensure safe operation under uncertainty.
For example, \cite{wei-qingFlexibilityEvaluationFlexible2012} incorporated penalty terms that reward higher flexibility indices into the objective function of an AC optimal power flow problem, which they heuristically solve by applying the Newton method.
\cite{bucherQuantificationFlexibilityPower2015} identified an optimal reserve procurement strategy for a transmission system operator under uncertainty using a linearized AC power flow model.
Other studies focused on maximizing the flexibility index, e.g.,    \cite{zhaoUnifiedFrameworkDefining2016} for the operation of a transmission system under nodal power injection uncertainty and \cite{leeRobustACOptimal2021} for a convex restriction of the AC optimal power flow problem.
\cite{gomezOperationalFlexibilityPower2019} applied the volumetric flexibility index \citep{laiProcessFlexibilityMultivariable2008} to approximate the secure operating region, considering N-1 contingency reliability by using linear sensitivity factors.
\cite{capitanescuPowerSystemFlexibility2021} determined the flexibility index of a power system using a spherical admissible uncertainty set and an AC power flow model.
More recently, \cite{zinglerOptimizingFlexibilityPower2025} maximized the region of manageable nodal injection uncertainties using a rigorous adaptive discretization-based approach for a linearized AC optimal power flow problem.

To construct the SCUC problem, we use the PyPSA-Eur framework \citep{horschPyPSAEurOpenOptimisation2018} and employ the k-means clustering method to obtain a reduced-size $3$-bus version of the German power grid \citep{horschRoleSpatialScale2017}.
For larger numbers of buses, the trained normalizing flow models became prohibitively large, resulting in subproblems that, in some instances, could not be solved within a reasonable time limit.
Consequently, the conditional flexibility index could not be determined, which is why we limit the analysis to the $3$-bus case.
To reduce the number of variables, we aggregate all generator types in each node into two categories: renewable and conventional.
Furthermore, we limit ourselves to the major generator categories:
Renewable generators encompass solar power plants and various types of wind power plants, while conventional generators include coal-fired power plants, gas-fired power plants, and nuclear power plants.
To train the normalizing flows, we use hourly historical renewable energy generation data from $2013$-$2017$, and we hold out historical data from $2018$ to validate our approach.
The renewable generation is calculated using PyPSA-Eur and is based on weather data from SARAH3 \citep{https://doi.org/10.5676/eum_saf_cm/sarah/v003} and ERA5 \citep{hersbachera5globalreanalysis2020}.
Our uncertain parameters are the capacity factors of the aggregated renewable generators, which we calculate by dividing the realized renewable generation by the total installed renewable capacity in each node, for every time step of the historical data.
Furthermore, we scale and shift the historical data before training the normalizing flow to improve performance.
Specifically, we shift the data by $-0.5$ and then scale it by a factor of $6$, leading to data in the range $[-3, 3]$.
The data from $2013$-$2017$ is split into training and validation data, where \SI{15}{\%} is used for model validation.
For further details on model training, we refer to Section $1$ of the supplementary material.
When embedding the normalizing flow, we implement the inverse of the scaling and shifting to obtain correct capacity factors.
Finally, we assume power demands to be certain to reduce problem complexity.
In principle, an extension to uncertain demands is possible by incorporating them as uncertain variables into the normalizing flow training.

\begin{figure}[H]
    \centering
    \includegraphics[width=0.5 \columnwidth]{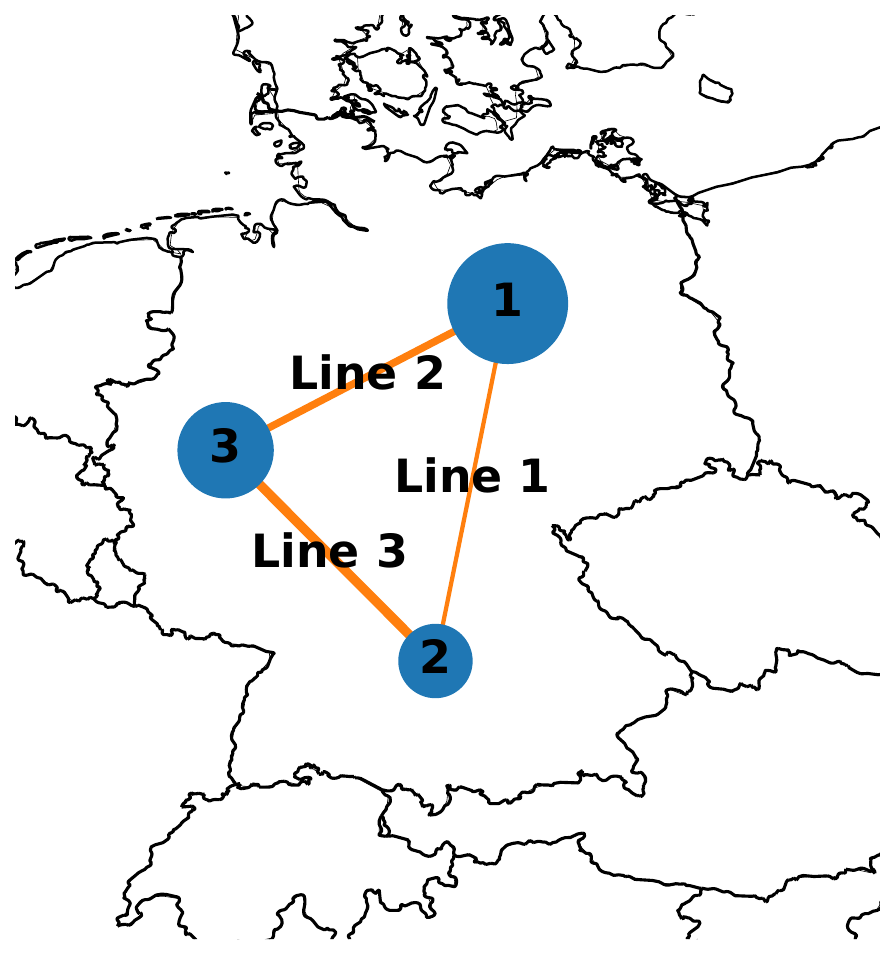}
    \caption{Topology of a reduced-size German power grid. Buses are shown as blue circles with their respective numbers; the size of the circle indicates the installed renewable capacity. Similarly, lines are shown in orange and line thickness indicates transmission capacity.}
    \label{fig:Topology}
\end{figure}

Figure \ref{fig:Topology} shows the topology of a reduced-size German power grid.
The size of the buses indicates the installed renewable capacity, while the thickness of the lines indicates transmission capacity.
We refer to Section $3$ of the supplementary material for detailed parameter values.

\subsection{Power system model}\label{sec:PSModel}
We model the transmission system behavior based on \cite{zinglerOptimizingFlexibilityPower2025} using a  linearized AC power flow approximation.
To keep this article self-contained, the main assumptions and governing equations are summarized below.

Decision variables $\mathbf{x}$ are the generator setpoints $P_{set, n}$ and the upper and lower bounds of generator power output $P_{ub, n}$ and $P_{lb, n}$ at each node $n \in \mathcal{N}$:
\begin{align*}
    \mathbf{x} = [&P_{set, 1}, \dots,  P_{set, |\mathcal{N}|}, P_{ub, 1}, \dots,  P_{ub, |\mathcal{N}|}, \\
    & P_{lb, 1}, \dots,  P_{lb, |\mathcal{N}|}]
\end{align*}
The generator setpoints are the only degrees of freedom; combined with the system constraints, they uniquely determine the corresponding upper and lower generation bounds.

The uncertain variables $\mathbf{y} = \left[P_{ren, 1}, \dots,  P_{ren, |\mathcal{N}|}, l_{1}, \dots,  l_{|\mathcal{N}|}\right]$ consist of renewable uncertainty realizations $P_{ren, n}$ and the normalizing flow latent variables $l_{n}$ for each node $n \in \mathcal{N}$.
The latent variables are the only degrees of freedom for the uncertain quantities and uniquely determine the renewable realizations according to
\begin{align*}
    P_{ren, n} = \max\left\{0, \mathbf{f}\left(\mathbf{l}, \mathbf{c}\right)_{n}\right\} P_{ren, nom, n} \ \forall n \in \mathcal{N},
\end{align*}
where $\mathbf{c}$ is the contextual information, $\mathbf{f}\left(\mathbf{l}, \mathbf{c}\right)_{n}$ is the $n$-th output of the normalizing flow transformation, and $P_{ren, nom, n}$ is the installed renewable capacity at node $n$.
The maximum operator ensures nonnegative renewable injections, correcting possible negative predictions caused by model errors.

To evaluate the impact of contextual information on CFI performance, we investigate different choices of $\mathbf{c}$.
First, we consider capacity factors from the previous time step, i.e., $\mathbf{c}_{t} = \mathbf{cf}_{t - 1}$.
Improved contextual information could, in principle, be obtained via forecasting models, although this is beyond the scope of the present study.
Second, we examine the effect of incorporating temporal features by including sine and cosine encodings of the hour of the day and the day of the year in the contextual information $\mathbf{c}$.

Contrary to \cite{zinglerOptimizingFlexibilityPower2025}, we do not have control variables; instead, the operational variables
\begin{align*}
    \mathbf{z} = [& P_{line,1}, \dots, P_{line,|\mathcal{L}_{line}|}, P_{gen,1}, \dots, P_{gen,|\mathcal{N}|}, \\
    & \theta_{1}, \dots, \theta_{|\mathcal{N}|}, \Delta_{inj} ]
\end{align*}
consist solely of implicit state variables that are uniquely determined by the decision variables, uncertainty realizations, and transmission system equations.
These variables include the power flows $P_{line, li}$ over each line $li \in \mathcal{L}_{line}$, the generator power outputs $P_{gen, n}$ and the voltage angles $\theta_n$ at each node $n \in \mathcal{N}$, and an artificially increased injection demand $\Delta_{inj, inc}$ that is introduced to model the generator behavior following \cite{zinglerOptimizingFlexibilityPower2025}.

The deviation of the generator outputs $P_{gen, n}$ from their setpoints $P_{set, n}$ to compensate for uncertain renewable generation $P_{ren, n}$ is determined as
\begin{align}\label{eq:GenControl}
    P_{gen, n} = \midop\left(P_{lb, n}, P_{set, n} + c_{g, n} \Delta_{inj, inc}, P_{ub, n}\right) \ \forall n \in \mathcal{N}.
\end{align}
The mid-operator ensures that the output $P_{gen, n}$ stays within the operating window, i.e., upper and lower bounds.
Uncertain renewable generation can lead to an \emph{injection demand}
\begin{align*}
    \Delta_{inj} = \underset{n \in \mathcal{N}}{\sum}\left(P_{dem, n} - P_{ren, n} - P_{set, n}\right),
\end{align*}
representing the difference between total demand and the combined injections from renewable and conventional generators at their setpoints.
To maintain overall power balance, the injection demand $\Delta_{inj}$ is allocated among the generators according to the participation factors $c_{g, n}$.
However, when a generator reaches its upper or lower operating limit, it can no longer contribute its assigned share.
To address this, an artificially increased injection demand $\Delta_{inj, inc}$ is introduced, redistributing the unmet demand from saturated generators among the remaining ones, consequently ensuring that the actual injection demand is fully met.

We assume that conventional generators can adjust their load up or down by \SI{5}{\%} of their rated capacity from the nominal operating point to react to uncertain renewable injections.
However, they must stay within their rated bounds $P_{ub, n}$ and $P_{lb, n}$ which are defined as
\begin{equation*}
    \begin{aligned}
        &P_{ub, n} = \min\left\{P_{set, n} + 0.05 P_{max, n}, P_{max, n} \right\}, \\
        &P_{lb, n} = \max\left\{0, P_{set, n} - 0.05 P_{max, n}\right\},
    \end{aligned}
\end{equation*}
where $P_{max, n}$ is the rated capacity.

To ensure safe operation, we impose that the rated line capacities $P_{nom, li}$ may not be exceeded: $\left|P_{line, li}\right| \le  P_{nom, li} \ \forall li \in \mathcal{L}_{line}$.
Since this is a vector-valued constraint, we use the formulation mentioned in Section \ref{sec:FIP} and obtain the GSIP-objective function:
\begin{align*}
    \max \bigl\{\lvert P_{line, li} \rvert - P_{nom, li} \ | \ li \in \mathcal{L}_{line}\bigr\} \le 0
\end{align*}
\cite{zinglerOptimizingFlexibilityPower2025} assume that the  overall power balance can be satisfied no matter the realization of the renewable generation.
Formally, this is expressed as
\begin{equation}
    \underset{n \in \mathcal{N}}{\sum} P_{lb, n} \le \underset{n \in \mathcal{N}}{\sum}\left(P_{dem, n} - P_{ren, n}\right) \le \underset{n \in \mathcal{N}}{\sum} P_{ub, n}.\label{eq:feas}
\end{equation}
In other words, the total power demand on the generators, i.e., the sum of all nodal power demands minus the sum of all nodal renewable injections, must lie between the sum of the lower bounds and the sum of the upper bounds of the generator power outputs.
However, we noticed in preliminary studies that this is not always the case in the German power grid example; the high renewable capacities create uncertainty scenarios in which net demand falls outside the feasible generation range, i.e., there is over- or under-production of power.
These scenarios render the power-balance constraint infeasible and therefore cannot be selected by the optimizer, which leads to erroneous estimates of flexibility.

To alleviate this issue, we include the constraints in Inequality \eqref{eq:feas}, i.e.,
\begin{equation*}
    \begin{aligned}
        &\underset{n \in \mathcal{N}}{\sum}\left(P_{lb, n} -  P_{dem, n} + P_{ren, n}\right) \le 0 \\
        &\underset{n \in \mathcal{N}}{\sum}\left(P_{dem, n} - P_{ren, n} - P_{ub, n}\right) \le 0, \\
    \end{aligned}
\end{equation*}
into the GSIP-objective function:
\begin{align*}
    g(\mathbf{x}, \mathbf{y}, \mathbf{z}) 
    &= \max \Biggl\{ &&\underset{n \in \mathcal{N}}{\sum} \bigl(P_{lb, n} - P_{dem, n} + P_{ren, n}\bigr), \\
    &&&\underset{n \in \mathcal{N}}{\sum} \bigl(P_{dem, n} - P_{ren, n} - P_{ub, n}\bigr), \\
    &&&\max \bigl\{\lvert P_{line, li} \rvert - P_{nom, li} \ | \ li \in \mathcal{L}_{line}\bigr\}\Biggr\} \le 0.
\end{align*}
If Inequality \eqref{eq:feas} is violated, the GSIP-objective function will be positive, indicating infeasibility of the overall GSIP.
On the contrary, any GSIP feasible solution will satisfy Inequality \eqref{eq:feas}.

Note that if Inequality \eqref{eq:feas} is not fulfilled, the system of equations to determine the operational variables cannot be solved as the overall power balance cannot be satisfied.
Consequently, line flows are undefined, and the satisfaction of line rating constraints cannot be evaluated.
To address this, we model the transmission system constraints as a generalized disjunctive program \citep{grossmannGeneralizedDisjunctiveProgramming2012} and only consider the rated line capacity constraints if Inequality \eqref{eq:feas} is fulfilled.
Other constraints that are enforced solely if Inequality \eqref{eq:feas} is fulfilled include the constraints describing the generator control, i.e., Equation \eqref{eq:GenControl}, the nodal power balances
\begin{align*}
    &P_{dem, n} - P_{ren, n} - P_{gen, n}\\
    & - \underset{li \in \mathcal{L}_{in}(n)}{\sum} P_{line, li} + \underset{li \in \mathcal{L}_{out}(n)}{\sum} P_{line, li} = 0 \ \forall n \in \mathcal{N},
\end{align*}
where $\mathcal{L}_{in}(n)$ and $\mathcal{L}_{out}(n)$ are the index sets of incoming and outgoing lines, and finally, the constraints determining the line flow, i.e.,
\begin{align*}
    P_{line, li} = h_{li}\left(\theta_{t(li)} - \theta_{o(li)}\right) \ \forall li \in \mathcal{L}_{line},
\end{align*}
where $h_{li}$ is the line admittance and $t(li)$ and $o(li)$ are the terminating and originating bus indices for line $li$.

Lastly, we form the SIP relaxation to resolve the dependency  of the lower-level feasible set on $\delta$ \citep{mitsosglobaloptimizationgeneralized2015} and obtain the existence-constrained semi-infinite objective function \citep{djelassidiskretisierungsbasiertealgorithmenfur2020}
\begin{equation*}
    \underset{\mathbf{y} \in \mathcal{Y}_{SIP}(\mathbf{c})}\max \underset{\mathbf{z} \in \mathcal{Z}(\mathbf{x}, \mathbf{y})}\min \min\bigl\{g(\mathbf{x}, \mathbf{y}, \mathbf{z}), \alpha \left(\delta - \left\lVert \mathbf{l} \right\rVert_{2}^2\right)\bigr\} \le 0,
\end{equation*}
where $\mathbf{l}$ are the normalizing flow latent variables contained in $\mathbf{y}$.
The parameter $\alpha$ is introduced to bring the GSIP objective function $g(\mathbf{x}, \mathbf{y}, \mathbf{z})$ and the constraint enforcing membership of the admissible uncertainty set to a similar scale, and is chosen as $\alpha = 500$.
For a detailed investigation of the influence of $\alpha$ on the solution time of similar problems, we refer to \cite{zinglerOptimizingFlexibilityPower2025} Section $8.2.1$.
Since the operational variables $\mathbf{z}$ are uniquely determined by $\mathbf{x}$ and $\mathbf{y}$ \citep{zinglerOptimizingFlexibilityPower2025}, we can move them into the maximization problem:
\begin{equation*}
    \underset{\mathbf{y} \in \mathcal{Y}_{SIP}(\mathbf{c}), \mathbf{z} \in \mathcal{Z}(\mathbf{x}, \mathbf{y})}\max \min\bigl\{g(\mathbf{x}, \mathbf{y}, \mathbf{z}), \alpha \left(\delta - \left\lVert \mathbf{l} \right\rVert_{2}^2\right)\bigr\}
\end{equation*}
We obtain the overall flexibility maximization problem:
\begin{equation}
\begin{aligned}\label{eq:SCUC}
    \delta_{cond}^{*}(\mathbf{c}) = &\underset{\delta \in \mathbb{R}, \mathbf{x} \in \mathcal{X}}\max \quad &\delta & \\
    &\text{s.t.} && P_{ub, n} = \min\left\{P_{set, n} + 0.05 P_{max, n}, P_{max, n} \right\} \\
    &&&P_{lb, n} = \max\left\{0, P_{set, n} - 0.05 P_{max, n}\right\}\\    
    &&& \underset{\mathbf{y} \in \mathcal{Y}_{SIP}(\mathbf{c}), \mathbf{z} \in \mathcal{Z}(\mathbf{x}, \mathbf{y})}\max \min\bigl\{g(\mathbf{x}, \mathbf{y}, \mathbf{z}), \alpha \left(\delta - \left\lVert \mathbf{l} \right\rVert_{2}^2\right)\bigr\} \le 0
\end{aligned}
\end{equation}
Additionally, we introduce the hypercube approach as a baseline method against which we compare the normalizing flow-based approach.
The hypercube problem differs from Problem \eqref{eq:CFI} in the definition of the admissible uncertainty set and leads to the embedded lower-level objective function
\begin{align*}
    \underset{\mathbf{y} \in \mathcal{Y}_{SIP}, \mathbf{z} \in \mathcal{Z}(\mathbf{x}, \mathbf{y})}\max \min\bigl\{g(\mathbf{x}, \mathbf{y}, \mathbf{z}), \alpha_{cube} \left( \delta - \left\lVert \mathbf{cf}_{cube}  - \mathbf{c}_{cube}\right\rVert_{\infty} \right)\bigr\} \le 0,
\end{align*}
where $\mathbf{cf}_{cube}$ is the capacity factor and $\mathbf{c}_{cube}$, i.e., the contextual information, is the center of the hypercube admissible uncertainty set.
The renewable generation in the lower-level problem is then calculated as:
\begin{align*}
    P_{ren, n} = cf_{cube, n} P_{ren, nom, n} \ \forall n \in \mathcal{N}
\end{align*}
For the detailed formulation of the embedded maxmin problems, of the CFI and the hypercube problems, we refer to Section $3$ of the supplementary materials.

\subsection{SCUC results}
First, we evaluate the quality of the normalizing flow models that are later embedded in the CFI problem.
Three models trained with different contextual information are compared.
The base model (NF) uses only capacity factor realizations from the previous time step as contextual input.
The second model (NF + T) additionally includes sine and cosine encodings of the hour of the day, while the third model (NF + T\&D) also incorporates sine and cosine encodings of the day of the year.
To control for differences in model capacity, all models use the same architecture: four RealNVP blocks, each parameterized by an MLP with one hidden layer of twelve units.
Again, an adaptive-discretization based algorithm \citep{blankenshipinfinitelyconstrainedoptimization1976} is applied, and the Gurobi solver \citep{gurobi} version $12.0.3$ is used to solve the subproblems.
Similar to the ``Two Moons'' example, the most crucial parameter was `NumericFocus', which was set to $2$ for the normalizing flows-based problems and $3$ for the hypercube-based problems to avoid erroneous results.
The remaining non-default solver parameters did not result in significant performance improvements in our experiments.
The feasibility tolerance is set to $25$ for the embedded lower-level problem, i.e., if the lower-level problem objective $\underset{\mathbf{y} \in \mathcal{Y}_{SIP}(\mathbf{c}), \mathbf{z} \in \mathcal{Z}(\mathbf{x}, \mathbf{y})}\max \min\bigl\{g(\mathbf{x}, \mathbf{y}, \mathbf{z}), \alpha \left(\delta - \left\lVert \mathbf{l} \right\rVert_{2}^2\right)\bigr\} \le 25$, the upper-level solution variables $\mathbf{x}$ and $\delta$ are considered feasible.
For comparison, renewable generation and demands, as well as line capacities, are in the order of \SI{1e5}{\mega\watt}.
The same feasibility tolerance applies to the hypercube problem.

In preliminary experiments, we noticed that if $\alpha_{cube}$ is chosen too small, the hypercube admissible uncertainty set contained points with significant infeasibility in some cases.
This occurs because the lower-level problem remains feasible at the edge of the admissible uncertainty set, as the term $\alpha_{cube} \left(\delta - \left\lVert \mathbf{cf}_{cube}  - \mathbf{c}_{cube}\right\rVert_{\infty}\right) \le 25$ holds regardless of the value of $g(\mathbf{x}, \mathbf{y}, \mathbf{z})$.
To mitigate this issue, we set $\alpha_{cube} = 50000$.

This large value ensures that the region of guaranteed feasibility near the boundary of the admissible uncertainty set becomes very narrow, thereby preventing the inclusion of significantly infeasible points, i.e., points for which $g(\mathbf{x}, \mathbf{y}, \mathbf{z}) \gg 25$.

\begin{figure}[H]
    \centering
    \includegraphics[width=0.75 \columnwidth]{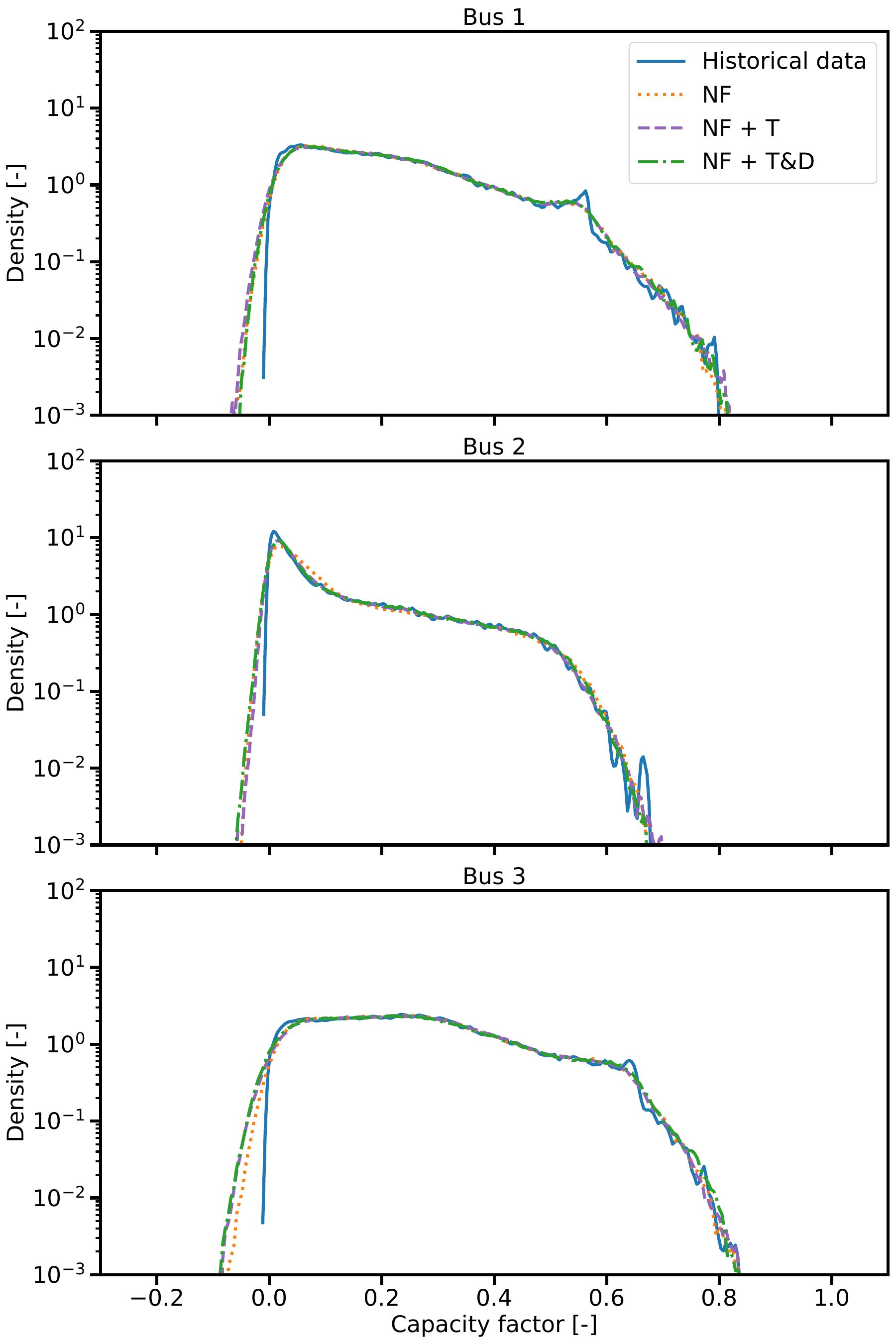}
    \caption{Density distributions of the historical data and samples drawn from the three normalizing flow models for each of the three buses. The base model (NF) uses the capacity factor realization in the previous timestep as contextual information; NF + T additionally uses the hour of the day, and NF + T\&D adds the day of the year. All three models approximate the data distribution well.}
    \label{fig:NF}
\end{figure}
Figure \ref{fig:NF} compares the empirical data distribution with samples from the trained normalizing flow models for each of the three buses.
For each timestep in the historical dataset, ten samples are drawn from each model.
All three normalizing flow models provide an accurate approximation of the empirical density of the data, with no significant performance differences.
However, at the lower tail of the distribution, the models generate unphysical negative capacity factors.
As described in Section \ref{sec:PSModel}, we address this by clipping negative values to zero, thereby ensuring physically consistent renewable generation realizations.

Next, we compare the results of solving Problem \eqref{eq:SCUC} with the embedded base model to those obtained using a hypercube admissible uncertainty set centered at the capacity factors of the previous time step.
Both approaches therefore use identical contextual information.

A small fraction of the validation dataset is inherently infeasible.
There are two causes for infeasibility, the most frequent cause with \SI{3.1}{\%} of cases is overproduction of renewables.
We do not consider curtailment, as it would be ideal in a robustness sense to curtail all renewable production to remove any uncertainty.
Hence, there are cases where total available renewable electricity exceeds demand and causes Problem \eqref{eq:SCUC} to become infeasible.
The other cause for infeasibility is underproduction in \SI{0.3}{\%} of cases, where there is barely any renewable production, and the remaining demand exceeds the total capacity of conventional generation.
In practice, these situations could be mitigated through storage technologies or cross-border electricity trading, both of which are excluded from the current analysis.

On average, the CFI achieves a conditional coverage of \SI{25}{\%}.
However, when evaluating operational feasibility against true uncertainty realizations, both the CFI and hypercube approaches yield feasible schedules in \SI{71}{\%} of cases.
The discrepancy between the conditional coverage and the share of feasible schedules indicates that the normalizing flow does not perfectly capture the conditional data distribution, despite accurately modeling the unconditional distribution (cf. Figure \ref{fig:NF}).

\begin{figure}[H]
    \centering
    \includegraphics[width=0.5 \columnwidth]{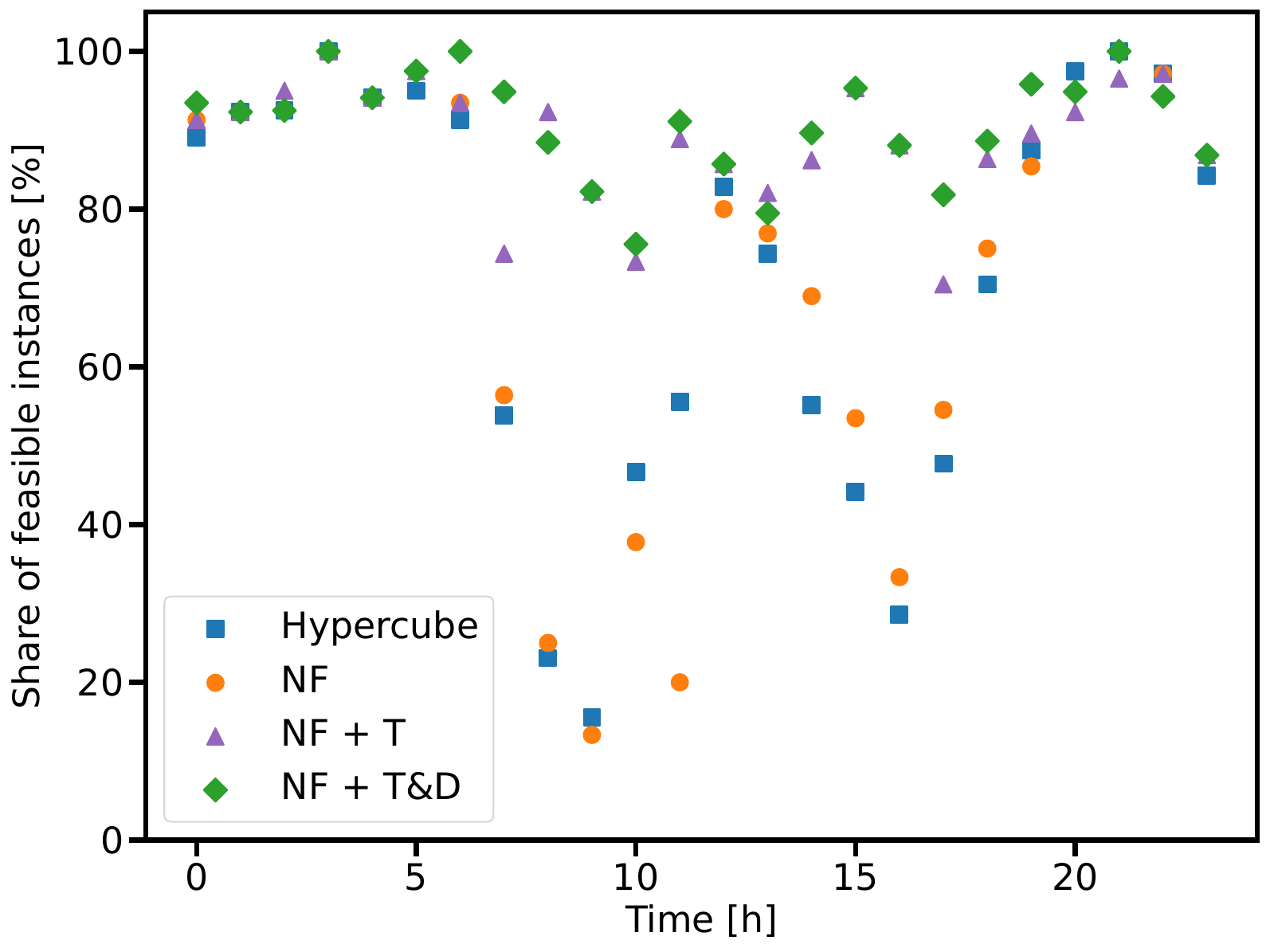}
    \caption{Share of feasible instances over the day for scheduling decisions determined by using hypercube, normalizing flow (NF), normalizing flow with time encoding (NF + T), and normalizing flow with time and day encoding (NF + T\&D) admissible uncertainty sets.}
    \label{fig:time_feas}
\end{figure}

To further investigate this, we compute scheduling decisions using all three normalizing flow models and the hypercube approach.
Figure \ref{fig:time_feas} shows the share of feasible instances by the hour of the day.
Both the base NF and hypercube approaches exhibit reduced feasibility in the morning ($5$ - $11$ a.m.) and afternoon ($12$ - $6$ p.m.), likely due to solar ramping periods with rapidly changing renewable output.
This supports the hypothesis that the base model fails to capture relevant temporal dependencies in the conditional distribution.

In contrast, models that include time-of-day or seasonal encodings generate schedules that have a higher and more stable share of feasible operation throughout the day.
The NF + T model generates schedules that are feasible in \SI{89}{\%} of cases, while the NF + T\&D model is the best and generates schedules that are feasible in \SI{91}{\%} of cases.
This demonstrates that suitable contextual information is crucial for practical applications.
Although all models approximate the unconditional data distribution well, incorporating temporal context substantially improves task-specific performance.

For the best-performing model (NF + T\&D), the average estimated coverage probability is \SI{56}{\%}, while \SI{91}{\%} of actual realizations are feasible.
As discussed in Section \ref{sec:NFUC}, the estimated coverage probability provides a lower bound on the true feasibility probability; however, this example illustrates that the gap between the two can be considerable.

\begin{figure}[H]
    \centering
    \includegraphics[width=0.5 \columnwidth]{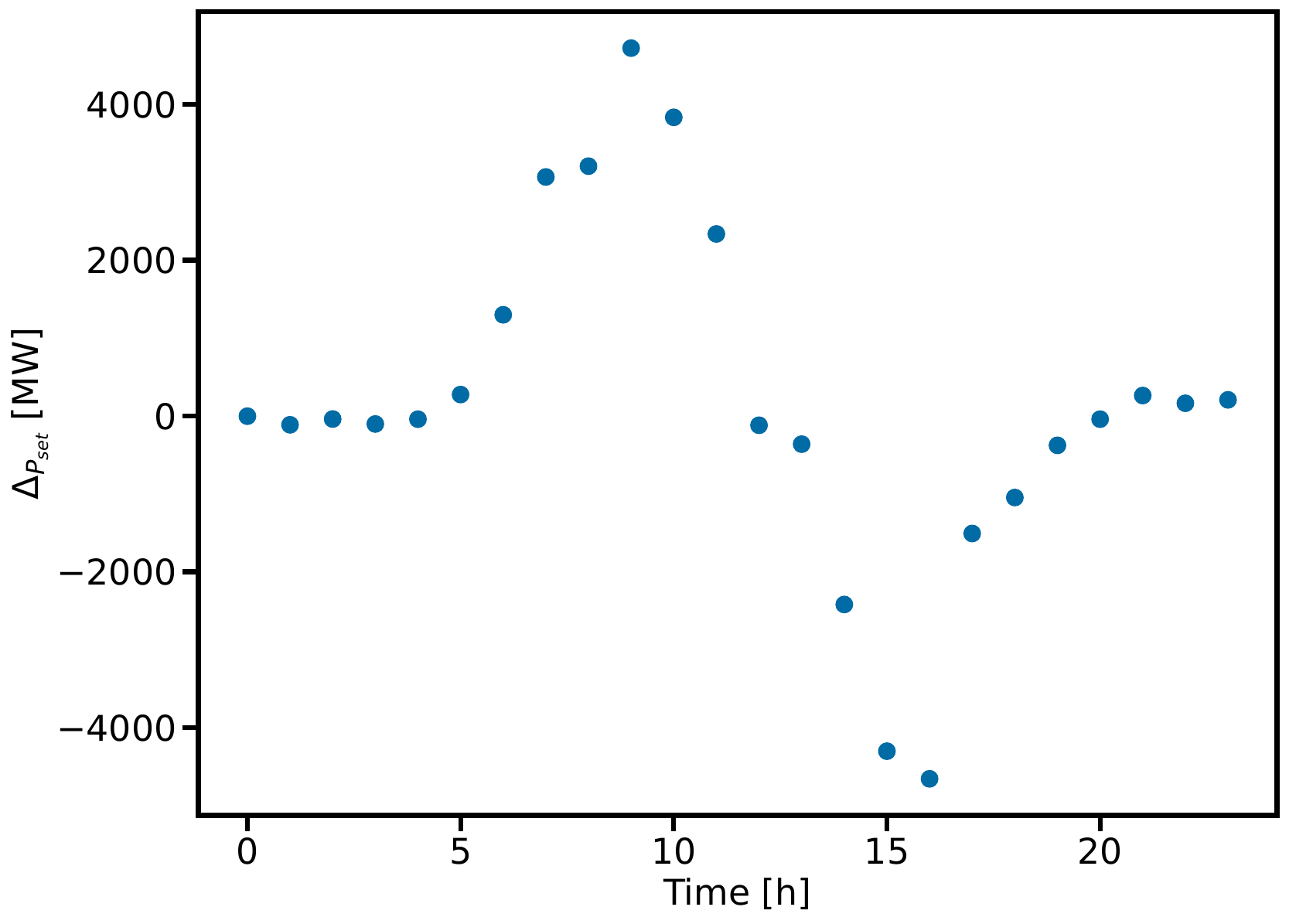}
    \caption{$\Delta_{P_{set}}(t) = \frac{1}{|\mathcal{D}(t)|}\underset{d \in \mathcal{D}(t)}{\sum}\underset{n \in \mathcal{N}}{\sum}\left(P_{set, n, cube, d} - P_{set, n, NF + T\&D, d}\right)$, average difference between the hypercube and NF + T\&D generator setpoints in each bus $n$ summed over all buses $\mathcal{N}$, plotted against the hour of the day. The average is calculated over all datapoints $d$ in the testset $\mathcal{D}(t)$ for each hour of the day $t$.}
    \label{fig:P_set_diff}
\end{figure}

Figure \ref{fig:P_set_diff} shows the average sum of differences between the hypercube and NF + T\&D generator setpoints for every hour of the day $\Delta_{P_{set}}(t)$, which is defined as: 
\begin{align*}
    \Delta_{P_{set}}(t) = \frac{1}{|\mathcal{D}(t)|}\underset{d \in \mathcal{D}(t)}{\sum}\underset{n \in \mathcal{N}}{\sum}\left(P_{set, n, cube, d} - P_{set, n, NF + T\&D, d}\right)
\end{align*}
where $\mathcal{D}(t)$ is the index set of datapoints for time of day $t$. 
The normalizing flow tends to schedule lower generation in the morning, when renewable output is rising, and higher generation in the afternoon, when it is falling.

For comparison, we also solve the hypercube problem with the center at the predicted renewable capacity factors of the NF + T\&D model, i.e., $\mathbf{c}_{cube} = \mathbf{f}_{NF + T\&D}\left(\mathbf{0}, \mathbf{c}\right)$.
The setpoints determined by solving this problem achieve feasible operation in \SI{90}{\%} of test-set uncertainty realizations compared to \SI{91}{\%} for the NF T\&D model.
Thus, most of the improvement stems from the ability of the normalizing flow to predict expected renewable generation rather than from the specific shape of the admissible uncertainty set.

\begin{figure}[H]
    \centering
    \includegraphics[width=0.5 \columnwidth]{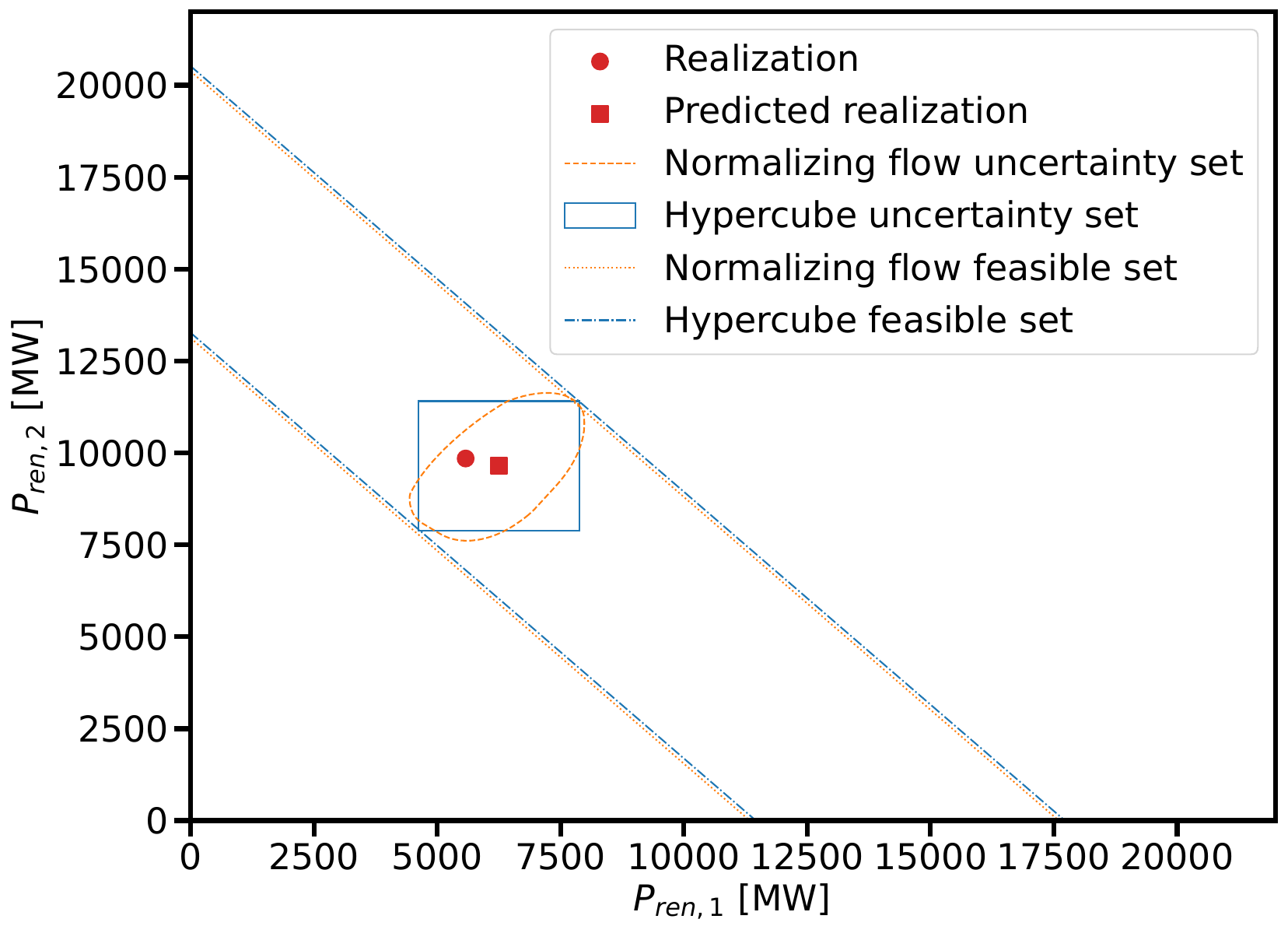}
    \caption{Uncertainty and feasible set for the normalizing flow (orange) and hypercube (blue) CFI approach for the $24$th of March $2018$ at $8$ a.m. on the $\mathit{2}$\emph{-bus} system. The true uncertainty realization is shown in red. Although the admissible uncertainty sets are different, they lead to similar scheduling decisions $\mathbf{x}$ and hence feasible sets.}
    \label{fig:SCUC_2}
\end{figure}
To gain further insight, we analyze a simplified $\mathit{2}$\emph{-bus} version of the SCUC case study, which allows for direct visualization of the results.
Figure \ref{fig:SCUC_2} shows the uncertainty and feasible sets for the normalizing flow- and hypercube-based CFI approaches, where the hypercube is centered at the realization predicted by the normalizing flow.
Although the admissible uncertainty sets differ in shape and they produce different generator setpoints ($\mathbf{P}_{set, NF} = \begin{bmatrix} 40490, & 1077 \end{bmatrix}^{T}$ for the normalizing flow and $\mathbf{P}_{set, cube} = \begin{bmatrix} 26145, & 15289 \end{bmatrix}^{T}$ for the hypercube), they lead to very similar admissible uncertainty region.
The boundaries of these admissible uncertainty regions correspond to lines of constant total renewable generation, indicating that total electricity production, rather than line capacity, constitutes the active constraint.
\begin{figure}[H]
    \centering
    \includegraphics[width=0.5 \columnwidth]{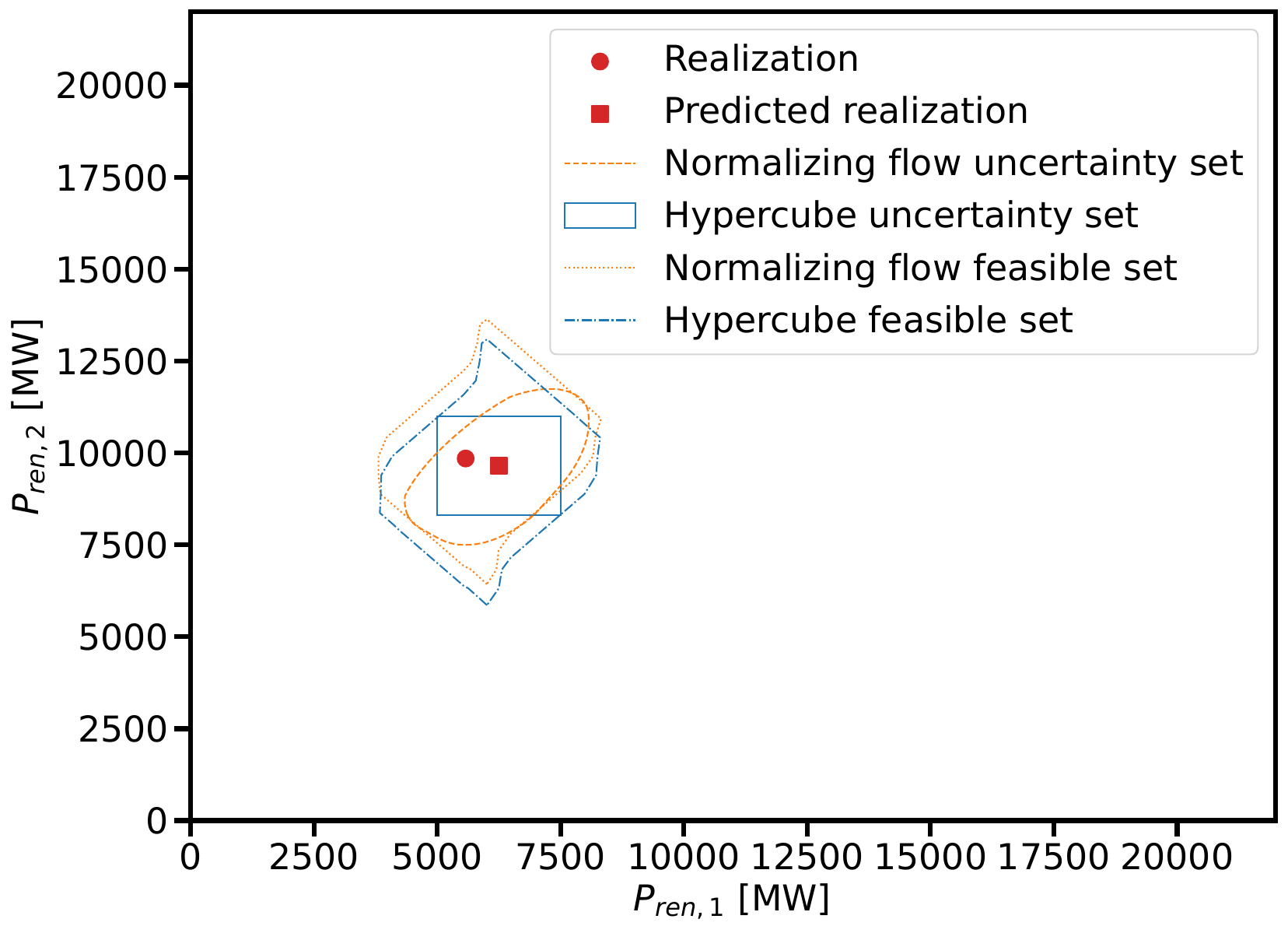}
    \caption{Uncertainty and feasible set for the normalizing flow (orange) and hypercube (blue) CFI approach for the $24$th of March $2018$ at $8$ a.m. on the $\mathit{2}$\emph{-bus} system with line capacities scaled by $0.05$. The true uncertainty realization is shown in red.}
    \label{fig:SCUC_2_rl}
\end{figure}
To test this hypothesis, we scale the line capacity by a factor of 0.05 and solve the problem again.
The resulting uncertainty and feasible sets are shown in Figure \ref{fig:SCUC_2_rl}.
The shape of the admissible uncertainty regions has changed significantly as line flow constraints now become active.
As a result, the generator set points have become more similar with $\mathbf{P}_{set, NF} = \begin{bmatrix} 26149, & 15221 \end{bmatrix}^{T}$ for the normalizing flow and $\mathbf{P}_{set, cube} = \begin{bmatrix} 26704, & 15170 \end{bmatrix}^{T}$ for the hypercube.
Although the scheduling decisions are now closer, the admissible uncertainty regions produced by the two approaches begin to diverge more noticeably.
Nevertheless, both methods achieve similar overall performance, with schedules remaining feasible in \SI{75}{\%} for the normalizing flow–based approach and \SI{76}{\%} for the hypercube–based approach.
Extending this analysis to the $3$-bus system with line capacities scaled by 0.2 yields comparable results: both the normalizing flow- and hypercube-based approaches produce feasible schedules in \SI{37}{\%} of cases.

In summary, for the SCUC problem with a strongly aggregated grid, the data distribution appears to be such that a flow-based admissible uncertainty set offers little advantage over a hypercube set if a reliable forecasting model provides a suitable conditional admissible uncertainty set center.
Nonetheless, the CFI approach can automatically learn appropriately centered conditional admissible uncertainty sets from historical data, whereas the hypercube approach requires a separate prediction model to determine the conditional uncertainty center.

\section{Conclusion}\label{sec:conclusion}
The growing availability of high-resolution data highlights the need for methods that can leverage contextual information to improve operational scheduling under uncertainty.
The flexibility index \citep{swaneyIndexOperationalFlexibility1985} can be applied to obtain robust schedules while accounting for correlations in the uncertain parameters \citep{rooneyDesignModelParameter2001}.
However, additional contextual information, such as historical observations beyond the directly uncertain variables, has not yet been considered, raising the question of whether incorporating contextual information could improve existing methods and provide more accurate estimates of system flexibility.

We present the conditional flexibility index (CFI), a variant of the flexibility index that constructs the admissible uncertainty set in the latent space of a normalizing flow (a generative machine learning model), allowing the consideration of uncertain parameter correlations and complex admissible uncertainty set shapes.
We solve the resulting existence-constrained generalized semi-infinite program using an adaptive discretization-based approach \citep{blankenshipinfinitelyconstrainedoptimization1976, mitsosglobaloptimizationgeneralized2015}.

The CFI can be applied to arbitrarily complex distributions of uncertain parameters to capture the shape of the admissible uncertainty set.
By leveraging the preservation of the probability measure of the transformation learned by the normalizing flow \citep{papamakarios2021normalizing}, an interpretation of the identified flexibility index as a lower bound to the probability of feasibility is possible.
However, embedding the normalizing flow results in a mixed-integer nonlinear program, for which the computational effort increases rapidly with the size of the normalizing flow.
Consequently, higher-dimensional uncertainty representations require greater model capacity, further amplifying computational demands.
Furthermore, due to the preservation of topological properties \citep{papamakarios2021normalizing}, the approximation quality of the normalizing flow suffers if the distribution of the uncertain parameters contains holes, i.e., is topologically different from the base distribution of the normalizing flow.

We demonstrate that no universal statement can be made regarding data-driven admissible uncertainty sets outperforming simple admissible uncertainty sets, nor conditional admissible uncertainty sets consistently outperforming unconditional ones.
This is because the centers of the admissible uncertainty sets are no longer user-specified nominal parameter values, but are instead determined from data or inferred using a forecasting model.
However, both data-driven and conditional admissible uncertainty sets have the advantage of excluding regions of the uncertain parameter space where no historical data exist, preventing these regions from limiting the flexibility index.
We apply the CFI to an illustrative example and a security-constrained unit commitment (SCUC) problem.
On the illustrative ``Two-Moons'' example, the conditional coverage probability of the normalizing flow-based admissible uncertainty sets is greater than that achieved by hypercubes.
Some variability is introduced by the training of the normalizing flow model.
However, this issue can be alleviated by training multiple models and choosing the one with the highest conditional coverage.

In the SCUC example, we compare normalizing flow models with access to different sets of contextual information.
All models can approximate the unconditional data distribution well; however, the normalizing flow models that have access to the hour of the day as contextual information generate higher-quality scheduling decisions.
The best normalizing flow model is the one provided with the richest contextual information — namely, the previous-timestep capacity factor realization, hour of the day, and day of the year — and yields schedules that are feasible for \SI{91}{\%} of test set realizations.
Its improved performance stems from accurately anticipating the characteristic morning and afternoon ramping of renewable generation, underscoring the importance of selecting informative contextual features for the CFI.


The hypercube admissible uncertainty set achieves performance similar to the normalizing flow set that does not have access to the temporal contextual information.
Moreover, when the hypercube admissible uncertainty set is centered at the renewable capacity factor predicted by the best normalizing flow model, the performance improves to \SI{90}{\%}, similar to the best flow model performance, highlighting the advantage of incorporating conditional information.
Importantly, in the CFI approach, the normalizing flow can directly learn the center of the admissible uncertainty set from the data, i.e., the normalizing flow learns how the capacity factor will change from the previous timestep to the current one, conditioned on the time of day, whereas the hypercube approach requires the practitioner to provide a suitable forecasting model to determine the center for the hypercube approach.

The applicability of the CFI to higher-dimensional problems can be enhanced by developing more effective algorithms for solving existence-constrained semi-infinite programs and refining existing mixed-integer nonlinear programming solvers.
Higher-dimensional problems may be solved by using heuristic solvers, e.g., genetic algorithms, to find approximate solutions to the involved subproblems; however, this comes at the cost of feasibility guarantees.
Furthermore, the coverage probability of conditional admissible uncertainty sets may be improved by treating the center of the admissible uncertainty set as an optimization variable or by removing the need to include the center at all and instead identifying a subset of the support of the data distribution that maximizes the coverage probability but does not need to contain the center at all.
However, it is presently unclear how such an uncertainty set could be constructed in practice.

\section*{Acknowledgment}
This work was performed as part of the Helmholtz School for Data Science in Life, Earth and Energy (HDS-LEE) and received funding from the Helmholtz Association of German Research Centres. We further acknowledge financial support by the Helmholtz Association of German Research Centres 
through program-oriented funding.

\section*{Data availability}
Data will be made available on request.

\section*{Declaration of competing interests}
Moritz Wedemeyer reports financial support was provided by Helmholtz Association of German Research Centres. Given his role as editor of Computers \& Chemical Engineering, Alexander Mitsos had no involvement in the peer review of this article and had no access to information regarding its peer review. Full responsibility for the editorial process for this article was delegated to another journal editor. If there are other authors, they declare that they have no known competing financial interests or personal relationships that could have appeared to influence the work reported in this paper.

\section*{Authors' contributions}
Conceptualization: M.W., E.C., A.M., M.D.; Methodology: M.W.; Software: M.W.; Formal analysis and investigation: M.W.; Visualization: M.W.; Writing - original draft preparation: M.W.; Writing - review and editing: E.C., A.M., M.D.; Funding acquisition: A.M., M.D.; Supervision: A.M., M.D.

\section*{Declaration of generative AI and AI-assisted technologies in the manuscript preparation process.}
During the preparation of this work, M.W. used ChatGPT and Grammarly in order to correct grammar and spelling and to improve the style of writing. After using these tools, all authors reviewed and edited the content as needed and take full responsibility for the content of the publication.

\section*{Nomenclature}
Throughout the manuscript, scalar-valued quantities are denoted in regular font, e.g., $x$, vector-valued quantities are denoted in bold font, e.g., $\mathbf{x}$, and set-valued quantities are denoted in calligraphic font, e.g., $\mathcal{X}$.
\subsection*{Abbreviations}
\begin{table}[H]
\begin{tabular}{ll}
CFI & Conditional flexibility index \\
EGSIP & Existence-constrained generalized semi-infinite program \\
GSIP & Generalized semi-infinite program \\
MLP & Multi-layer perceptron \\
NF & Normalizing flow \\
SCUC & Security-constrained unit commitment \\
SIP & Semi-infinite program \\
\end{tabular}
\end{table}

\subsection*{Greek Symbols}
\begin{table}[H]
\begin{tabular}{ll}
$\alpha$ & Scaling factor \\
$\delta$ & Flexibility index \\
$\delta_{cond}$ & Conditional flexibility index \\
$\Delta_{inj}$ & Injection demand \\
$\Delta_{inj, inc}$ & Artificially increased injection demand \\
$\Delta_{P_{set}}(t)$ & Average difference between generator setpoints at hour of the day $t$\\
$\theta$ & Voltage angle\\
\end{tabular}
\end{table}

\subsection*{Latin Symbols}
{\allowdisplaybreaks
\begin{table}[H]
\begin{tabular}{ll}
$\mathbf{A}$ & Selection matrix \\
$\mathbf{b}$ & Bias \\
$c$ & Contextual information \\
$\mathbf{c}$ & Vector of contextual information \\
$cf$ & Capacity factor realization \\
$c_{g}$ & Contribution factor \\
$\mathbf{cf}$ & Vector of capacity factor realizations \\
$\mathcal{D}(t)$ & Dataset for hour of the day $t$ \\
$\mathbf{f}(\mathbf{l}, \mathbf{c})$ & Normalizing flow transformation \\
$g$ & Constraint \\
$h(\mathbf{y})$ & Himmelblau function \\
$h_{li}$ & Line admittance [\si{\mega\watt}] \\
$i$ & Index \\
$j$ & Index \\
$\mathcal{J}$ & Set of indices \\
$k$ & Index \\
$l$ & Latent variable \\
$\mathbf{l}$ & Vector of latent variables \\
$\mathcal{L}$ & Feasible set of latent variables \\
$\mathcal{L}_{line}$ & Set of line indices \\
\end{tabular}
\end{table}
\begin{table}[H]
\begin{tabular}{ll}
$\mathcal{L}_{in}$ & Set of incoming line indices \\
$\mathcal{L}_{out}$ & Set of outgoing line indices \\
$li$ & Line index \\
$n$ & Node index \\
$\mathcal{N}$ & Set of node indices \\
$p_{l}(\mathbf{l})$ & Probability distribution of latent variables \\
$p_{y}(\mathbf{y})$ & Probability distribution of uncertain variables \\
$P_{dem}$ & Electricity demand [\si{\mega\watt}] \\
$P_{gen}$ & Generator output [\si{\mega\watt}] \\
$P_{lb}$ & Generator lower bound [\si{\mega\watt}] \\
$P_{line}$ & Line power flow [\si{\mega\watt}] \\ 
$P_{max}$ & Nominal generator capacity [\si{\mega\watt}] \\
$P_{nom}$ & Nominal line capacity [\si{\mega\watt}] \\
$P_{ren}$ & Renewable electricity generation [\si{\mega\watt}] \\
$P_{ren, nom}$ & Renewable generation nominal capacity [\si{\mega\watt}] \\
$P_{set}$ & Generator setpoint [\si{\mega\watt}] \\
$P_{ub}$ & Generator upper bound [\si{\mega\watt}] \\
$\mathbf{s}$ & Scaling vector \\
$\mathbf{t}$ & Translation vector \\
$\mathbf{x}$ & Vector of design variables \\
$\mathcal{X}$ & Feasible set of design variables \\
$y$ & Uncertainty variable \\
$\mathbf{y}$ & Vector of uncertainty variables \\
$\mathcal{Y}$ & Feasible set of uncertainty variables \\
$\mathcal{Y}_{SIP}$ & Feasible set of uncertainty variables for the SIP relaxation \\
$\mathcal{Y}_{cond}$ & Conditonal feasible set of uncertainty variables \\
$\mathbf{z}$ & Vector of operational variables \\
$\mathcal{Z}$ & Feasible set of operational variables \\
\end{tabular}
\end{table}

\subsection*{Subscripts}
\begin{table}[H]
\begin{tabular}{ll}
$cube$ & Hypercube \\
$d$ & Datapoint index\\
$i$ & Index \\
$j$ & Index \\
$li$ & Line index \\
$n$ & Node index \\
$o(li)$ & Originating index of line $li$\\
$t(li)$ & Terminating index of line $li$\\
\end{tabular}
\end{table}

\subsection*{Superscripts}
\begin{table}[H]
\begin{tabular}{ll}
$*$ & Optimal value \\
\end{tabular}
\end{table}
 
\bibliographystyle{elsarticle-harv}
  \renewcommand{\refname}{References}  
  \bibliography{bibliography.bib}
\end{document}


\thispagestyle{firststyle}

  \begin{center}
    \begin{large}
      \textbf{\mytitle}
    \end{large} \\
    \myauthor
  \end{center}

  \vspace{0.5cm}

  \begin{footnotesize}
    \affil
  \end{footnotesize}

  \vspace{0.5cm}
\section{Normalizing flow training}
In this section, we describe the training settings for the normalizing flow.
We implement and train the models using PyTorch Lightning \citep{Falcon_PyTorch_Lightning_2019}.
We use the Adam optimizer \citep{kingma2017adammethodstochasticoptimization} and a learning rate of $0.005$.
Furthermore, a learning rate scheduler that reduces the learning rate by $0.5$ if the validation loss does not decrease after $10$ epochs is used.
We terminate the training if the validation loss does not improve by $0.001$ in $20$ epochs (early stopping).
The parameters that achieved the best validation loss are then used as the final model.
For the ``Two-Moons'' and ``Annulus'' data sets, we use a gradient clipping value of $0.5$ and a batch size of $8192$, and for the SCUC dataset, a gradient clipping value of $1000$ and a batch size of $2048$.
Although we did not see a strong influence of the gradient clipping value, we provide it here for completeness.

\section{Two-moons example}
\begin{figure}[H]
    \centering
    \includegraphics[width=200pt]{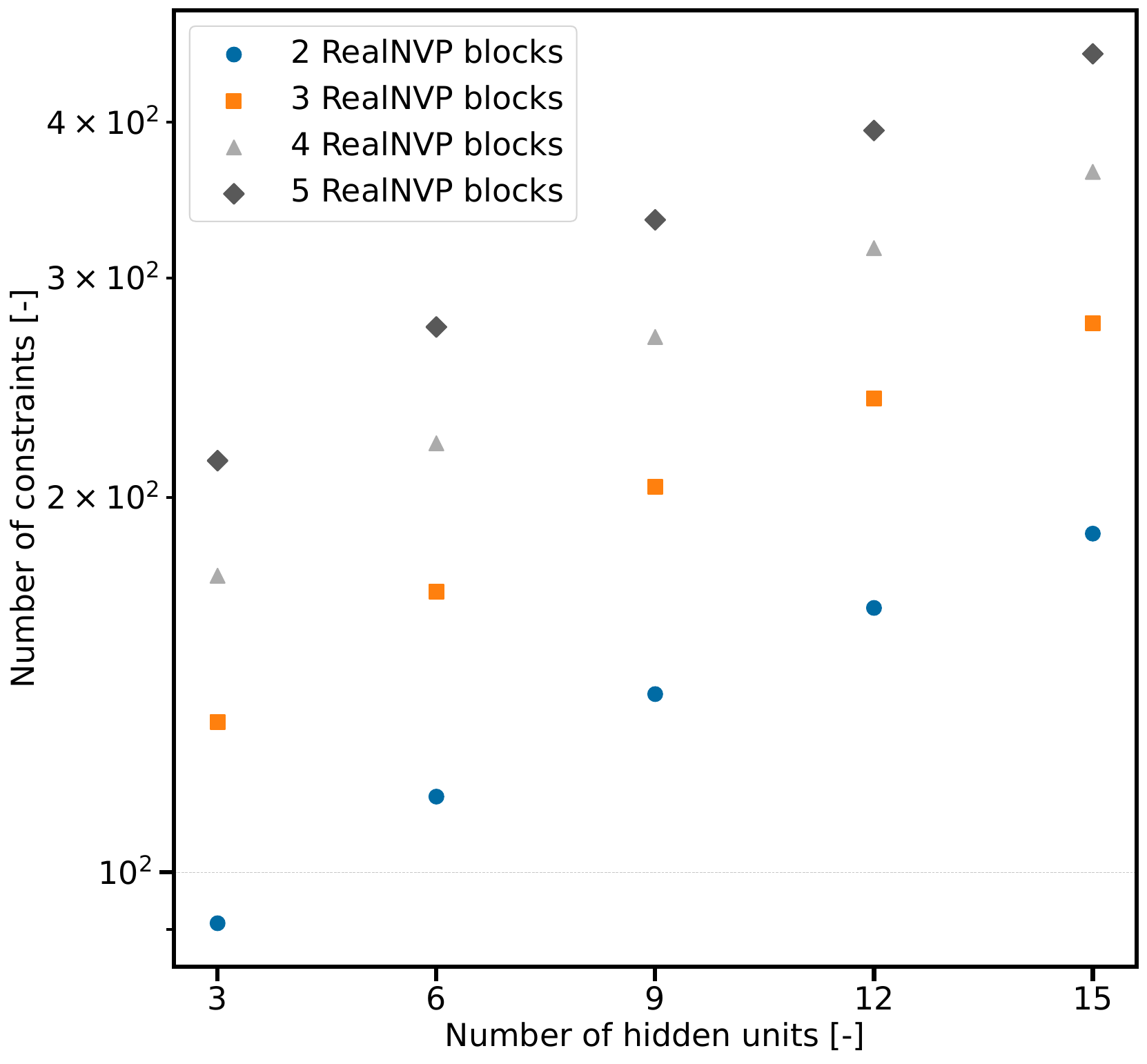}
    \caption{Number of constraints added to the CFI problem to embed the normalizing flow model for varying number of RealNVP blocks and hidden units.}
    \label{fig:Problemsize}
\end{figure}

Figure \ref{fig:Problemsize} shows the number of additional constraints required to embed the normalizing flow model into the CFI problem, which mirrors the trend in average runtime.
A similar increase is also evident in the number of variables, see Table \ref{tab:size}.

\begin{table}
    \begin{center}
        \begin{tabular}{c|c|c|c|c|c|c|c}
            \hline
            RealNVP & Hidden & Continuous & Integer & Linear & General & Quadratic & LLP\\
            blocks & units & variables &  variables & constraints &  constraints &  constraints & runtime [s]\\
            \hline
            \multirow{5}{4em}{2} & 3 & 100 & 4 & 55 & 30 & 6 & 0.21\\
            & 6 & 124 & 4 & 67& 42 & 6 & 0.33\\
            & 9 & 148 & 4 & 79 & 54 & 6 & 0.67\\
            & 12 & 172 & 4 & 91 & 66 & 6 & 0.78\\
            & 15 & 196 & 4 & 103 & 78 & 6 & 0.95\\
            \hline
            \multirow{5}{4em}{3} & 3 & 145 & 6 & 78 & 45 & 9 & 0.29\\
            & 6 & 181 & 6 & 96 & 63 & 9 & 0.76\\
            & 9 & 217 & 6 & 114 & 81 & 9 & 1.48\\
            & 12 & 253 & 6 & 132 & 99 & 9 & 2.24\\
            & 15 & 289 & 6 & 150 & 117 & 9 & 3.11\\
            \hline
            \multirow{5}{4em}{4} & 3 & 190 & 8 & 101 & 60 & 12 & 0.46\\
            & 6 & 238 & 8 & 125 & 84 & 12 & 1.56\\
            & 9 & 286 & 8 & 149 & 108 & 12 & 2.86\\
            & 12 & 334 & 8 & 173 & 132 & 12 & 4.64\\
            & 15 & 382 & 8 & 197 & 156 & 12 & 8.25\\
            \hline
            \multirow{5}{4em}{5} & 3 & 235 & 10 & 124 & 75 & 15 & 0.58\\
            & 6 & 295 & 10 & 154 & 105 & 15 & 2.43\\
            & 9 & 355 & 10 & 184 & 135 & 15 & 4.95\\
            & 12 & 415 & 10 & 214 & 165 & 15 & 11.83\\
            & 15 & 475 & 10 & 244 & 195 & 15 & 18.18\\
            \hline
        \end{tabular}
        \caption{Number of constraints and variables introduced to embed the normalizing flow model into the optimization problem \emph{before presolving} for different model sizes, parametrized by the number of RealNVP blocks and hidden units. Additionally, the average runtime reported by Gurobi version $12.0.3$ \citep{gurobi} to solve the lower-level problem (LLP) is given in the last column.}
        \label{tab:size}
    \end{center}
\end{table}

The Gurobi solver \citep{gurobi} used in this study distinguishes between different types of constraints:
Linear and quadratic constraints consist of linear and quadratic expressions, respectively, whereas general constraints include more complex, nonlinear expressions, for example, the max constraint $x_1 = \max \{x_2, x_3\}$.
Certain constraints, such as max constraints, are reformulated by the solver at solve time into a corresponding mixed-integer linear problem (MILP) formulation.
The total number of introduced linear constraints and integer variables varies depending on the bounds on the variables and the parameter values in the corresponding expressions.
Hence, the parameters of the normalizing flow model influence the model size after presolving.

For illustration, consider a lower-level problem with an embedded normalizing flow model with $3$ RealNVP blocks, $9$ hidden layers, and fixed context $c=0$.
The original lower-level problem has $4$ quadratic constraints to model the Himmelblau constraint and $9$ additional quadratic constraints introduced by the normalizing flow.
It also contains $81$ general constraints, $75$ of which can be reformulated in MILP form, and $6$ of which are nonconvex constraints to model the softplus operation.
In total, the model has $220$ continuous variables and $6$ integer variables.
For one random seed, the presolved model contains $135$ SOS constraints, $4$ quadratic constraints, $5$ bilinear constraints, and $6$ nonlinear constraints with $177$ continuous and $132$ integer variables.
For another random seed, corresponding to another normalizing flow model with the same hyperparameters, the presolved model contains $126$ SOS constraints, $4$ quadratic constraints, $5$ bilinear constraints, $6$ nonlinear constraints, $174$ continuous, and $123$ integer variables.

These examples illustrate that, while the introduced variables and constraints for a given normalizing flow model of fixed size can be precisely determined, the presolved model size varies depending on the normalizing flow parameter values.

\section{Security-constrained unit commitment}
Table \ref{tab:pars} lists the parameter values used in our study.
\begin{table}
    \begin{center}
        \begin{tabular}{c|c|c|c}
            \hline
            Parameter & Bus 1 & Bus 2 & Bus 3\\
            \hline
            Renewable capacity [\si{\mega\watt}] & 58421 & 22144 & 37179\\
            Conventional capacity [\si{\mega\watt}] & 19400 & 9886 & 38126\\
            \hline
            Parameter & Line 1 & Line 2 & Line 3\\
            \hline
            Transmission capacity [\si{\mega\watt}] & 10189 & 17741 & 22612\\
            Admittance [\si{\mega\watt}] & 8030 & 18653 & 24791\\
            \hline
        \end{tabular}
        \caption{Parameter values of the reduced-size three-bus German grid. Note that the admittance is also given in \si{\mega\watt}.}
        \label{tab:pars}
    \end{center}
\end{table}

Next, we give the problem formulation of the embedded MaxMin problem in the conditional flexibility index (CFI) problem (cf. Equation (6) in Section 4.1 of the main manuscript).
\begin{equation}
\begin{aligned}
    \phi(\mathbf{x}, \delta) = \underset{\mathbf{y} \in \mathcal{Y}, \mathbf{z} \in \mathcal{Z}}\max &\min\bigl\{g(\mathbf{x}, \mathbf{y}, \mathbf{z}), \alpha \left(\delta - \left\lVert \mathbf{l} \right\rVert_{2}^2 \right)\bigr\} \\
     &\text{s.t.} \begin{bmatrix} \underset{n \in \mathcal{N}}{\sum}\left(P_{lb, n} -  P_{dem, n} + P_{ren, n}\right) \le 0 \land \underset{n \in \mathcal{N}}{\sum}\left(P_{dem, n} - P_{ren, n} - P_{ub, n}\right) \le 0\\
      P_{gen, n} = \midop\left(P_{lb, n}, P_{set, n} + c_{g, n} \Delta_{inj, inc}, P_{ub, n}\right) \ \forall n \in \mathcal{N}\\
      P_{dem, n} - P_{ren, n} - P_{gen, n} - \underset{li \in \mathcal{L}_{in}(n)}{\sum} P_{line, li} + \underset{li \in \mathcal{L}_{out}(n)}{\sum} P_{line, li} = 0 \ \forall n \in \mathcal{N}\\
      P_{line, li} = h_{li}\left(\theta_{t(li)} - \theta_{o(li)}\right) \ \forall li \in \mathcal{L}_{line}
     \end{bmatrix} \lor  \\
    &\quad \ \begin{bmatrix} \neg \left( \underset{n \in \mathcal{N}}{\sum}\left(P_{lb, n} -  P_{dem, n} + P_{ren, n}\right) \le 0 \land \underset{n \in \mathcal{N}}{\sum}\left(P_{dem, n} - P_{ren, n} - P_{ub, n}\right) \le 0 \right)\\
    P_{line, li} = 0 \ \forall li \in \mathcal{L}_{line}
     \end{bmatrix}\\
      & \quad \ P_{ren, n} = \max\left\{0, \mathbf{f}\left(\mathbf{l}, \mathbf{c}\right)_{n}\right\} P_{ren, nom, n} \ \forall n \in \mathcal{N}
\end{aligned}
\end{equation}
with
\begin{align*}
    g(\mathbf{x}, \mathbf{y}, \mathbf{z}) 
    &= \max \Biggl\{ &&\underset{n \in \mathcal{N}}{\sum} \bigl(P_{lb, n} - P_{dem, n} + P_{ren, n}\bigr), \\
    &&&\underset{n \in \mathcal{N}}{\sum} \bigl(P_{dem, n} - P_{ren, n} - P_{ub, n}\bigr), \\
    &&&\underset{li \in \mathcal{L}_{line}}{\max} \bigl(\lvert P_{line, li} \rvert - P_{nom, li}\bigr)\Biggr\} \le 0
\end{align*}
and 
\begin{align*}
    \mathcal{Y} = \{\left(\mathbf{P}_{ren}, \mathbf{l}\right) \in \mathbb{R}_{\ge0}^{|\mathcal{N}|} \times \mathbb{R}^{|\mathcal{N}|} \} \\
    \mathcal{Z} = \{\left(\mathbf{P}_{gen}, \boldsymbol{\theta}, \mathbf{P}_{line}, \Delta_{inj, inc}\right) \in \mathbb{R}_{\ge0}^{|\mathcal{N}|} \times \mathbb{R}^{|\mathcal{N}|} \times \mathbb{R}^{|\mathcal{L}_{line}|} \times \mathbb{R}\}
\end{align*}

Leading to the SCUC CFI problem:
\begin{equation}
\begin{aligned}\label{eq:SCUC}
    \delta_{cond}^{*}(\mathbf{c}) = &\underset{\delta \in \mathbb{R}, \mathbf{x} \in \mathcal{X}}\max & \delta && \\
    &\text{s.t.} && P_{ub, n} = \min\left\{P_{set, n} + 0.05 P_{max, n}, P_{max, n} \right\} \forall n \in \mathcal{N} \\
    &&&P_{lb, n} = \max\left\{0, P_{set, n} - 0.05 P_{max, n}\right\}  \forall n \in \mathcal{N} \\    
    &&& \phi(\mathbf{x}, \delta) \le 0,
\end{aligned}
\end{equation}
with
\begin{align*}
    \mathcal{X} = \{\left(\mathbf{P}_{set}, \mathbf{P}_{ub}, \mathbf{P}_{lb}\right) \in \mathbb{R}_{\ge0}^{|\mathcal{N}|} \times \mathbb{R}_{\ge0}^{|\mathcal{N}|} \times \mathbb{R}_{\ge0}^{|\mathcal{N}|}\}.
\end{align*}

Next, we give the full formulation for the lower-level problem of the hypercube CFI problem:
\begin{equation}
\begin{aligned}
    \phi_{cube}(\mathbf{x}, \delta) = \underset{\mathbf{y} \in \mathcal{Y}_{cube}, \mathbf{z} \in \mathcal{Z}}\max &\min\big\{g(\mathbf{x}, \mathbf{y}, \mathbf{z}), \alpha_{cube} \left(\delta - \left\lVert \mathbf{cf}_{cube}  - \mathbf{c}_{cube}\right\rVert_{\infty} \right)\bigr\}\\
     &\text{s.t.} \begin{bmatrix} \underset{n \in \mathcal{N}}{\sum}\left(P_{lb, n} -  P_{dem, n} + P_{ren, n}\right) \le 0 \land \underset{n \in \mathcal{N}}{\sum}\left(P_{dem, n} - P_{ren, n} - P_{ub, n}\right) \le 0\\
      P_{gen, n} = \midop\left(P_{lb, n}, P_{set, n} + c_{g, n} \Delta_{inj, inc}, P_{ub, n}\right) \ \forall n \in \mathcal{N}\\
      P_{dem, n} - P_{ren, n} - P_{gen, n} - \underset{li \in \mathcal{L}_{in}(n)}{\sum} P_{line, li} + \underset{li \in \mathcal{L}_{out}(n)}{\sum} P_{line, li} = 0 \ \forall n \in \mathcal{N}\\
      P_{line, li} = h_{li}\left(\theta_{t(li)} - \theta_{o(li)}\right) \ \forall li \in \mathcal{L}_{line}
     \end{bmatrix} \lor  \\
    &\quad \ \begin{bmatrix} \neg \left( \underset{n \in \mathcal{N}}{\sum}\left(P_{lb, n} -  P_{dem, n} + P_{ren, n}\right) \le 0 \land \underset{n \in \mathcal{N}}{\sum}\left(P_{dem, n} - P_{ren, n} - P_{ub, n}\right) \le 0 \right)\\
    P_{line, li} = 0 \ \forall li \in \mathcal{L}_{line}
     \end{bmatrix}\\
      & \quad \ P_{ren, n} = cf_{cube, n} P_{ren, nom, n} \ \forall n \in \mathcal{N}
\end{aligned}
\end{equation}
with
\begin{align*}
    \mathcal{Y}_{cube} = \{\left(\mathbf{P}_{ren}, \mathbf{cf}_{cube}\right) \in \mathbb{R}_{\ge0}^{|\mathcal{N}|} \times [0, 1]^{|\mathcal{N}|} \} \\
\end{align*}

\section*{Nomenclature}
Throughout the manuscript, scalar-valued quantities are denoted in regular font, e.g., $x$, vector-valued quantities are denoted in bold font, e.g., $\mathbf{x}$, and set-valued quantities are denoted in calligraphic font, e.g., $\mathcal{X}$.
\subsection*{Abbreviations}
\begin{table}[H]
\begin{tabular}{ll}
CFI & Conditional flexibility index \\
SCUC & Security-constrained unit commitment \\
\end{tabular}
\end{table}

\subsection*{Greek symbols}
\begin{table}[H]
\begin{tabular}{ll}
$\alpha$ & Scaling factor \\
$\delta$ & Flexibility index \\
$\delta_{cond}$ & Conditional flexibility index \\
$\Delta_{inj, inc}$ & Artificially increased injection demand \\
$\phi(\mathbf{x}, \delta)$ & Optimal value function\\
$\theta$ & Voltage angle\\
\end{tabular}
\end{table}

\subsection*{Latin symbols}
\begin{table}[H]
\begin{tabular}{ll}
$c_{g}$ & Contribution factor \\
$\mathbf{c}$ & Vector of contextual information \\
$cf$ & Capacity factor realization \\
$\mathbf{cf}$ & Vector of capacity factor realizations \\
$\mathbf{f}(\mathbf{l}, \mathbf{c})$ & Normalizing flow transformation \\
$g$ & Constraint \\
$h_{li}$ & Line admittance [\si{\mega\watt}] \\
$l$ & Latent variable \\
$\mathbf{l}$ & Vector of latent variables \\
$\mathcal{L}$ & Feasible set of latent variables \\
$\mathcal{L}_{line}$ & Set of line indices \\
$\mathcal{L}_{in}$ & Set of incoming line indices \\
$\mathcal{L}_{out}$ & Set of outgoing line indices \\
$li$ & Line index \\
$n$ & Node index \\
$\mathcal{N}$ & Set of node indices \\
$P_{dem}$ & Electricity demand [\si{\mega\watt}] \\
$P_{gen}$ & Generator output [\si{\mega\watt}] \\
$P_{lb}$ & Generator lower bound [\si{\mega\watt}] \\
$P_{line}$ & Line power flow [\si{\mega\watt}] \\ 
$P_{max}$ & Nominal generator capacity [\si{\mega\watt}] \\
$P_{nom}$ & Nominal line capacity [\si{\mega\watt}] \\
$P_{ren}$ & Renewable electricity generation [\si{\mega\watt}] \\
$P_{ren, nom}$ & Renewable generation nominal capacity [\si{\mega\watt}] \\
$P_{set}$ & Generator setpoint [\si{\mega\watt}] \\
$P_{ub}$ & Generator upper bound [\si{\mega\watt}] \\
$\mathbf{x}$ & Vector of design variables \\
$\mathcal{X}$ & Feasible set of design variables \\
$\mathbf{y}$ & Vector of uncertainty variables \\
$\mathcal{Y}$ & Feasible set of uncertainty variables \\
$\mathbf{z}$ & Vector of operational variables \\
$\mathcal{Z}$ & Feasible set of operational variables \\
\end{tabular}
\end{table}

\subsection*{Subscripts}
\begin{table}[H]
\begin{tabular}{ll}
$cube$ & Hypercube \\
$li$ & Line index \\
$n$ & Node index \\
$o(li)$ & Originating index of line $li$\\
$t(li)$ & Terminating index of line $li$\\
\end{tabular}
\end{table}

\subsection*{Superscripts}
\begin{table}[H]
\begin{tabular}{ll}
$*$ & Optimal value \\
\end{tabular}
\end{table}

\bibliographystyle{elsarticle-harv}
  \renewcommand{\refname}{References}  
  \bibliography{bibliography.bib}